\documentclass[a4paper,11pt]{article}
\usepackage[utf8]{inputenc}

\usepackage{authblk}
\usepackage{array}
\usepackage{amsmath}

\usepackage{amssymb}

\usepackage{graphicx}

  \usepackage{soul, color}
  \sethlcolor{green}
  \usepackage{url}
 \usepackage[neveradjust]{paralist}
\usepackage[top=2.54cm, bottom=2.54cm, left=2.54cm, right=2.54cm]{geometry} 

\begin{document}

\title{KF metamodel formalization}
\author[1,2]{Pablo Rub\'en Fillottrani}
\affil[1]{Departamento de Ciencias e Ingenier\'ia de la Computaci\'on, 
Universidad Nacional del Sur, 
Bah\'ia Blanca, Argentina}
\affil[2]{Comisi\'on de Investigaciones Cient\'{\i}ficas, Provincia de Buenos Aires, Argentina}
\author[3]{C. Maria Keet}
\affil[3]{Department of Computer Science, University of Cape Town, South Africa}
\date{ }

\maketitle

\begin{abstract}
The KF metamodel \cite{KF13er,KF14dke} is a comprehensive unifying metamodel covering
the static structural entities and constraints of UML Class Diagrams (v2.4.1), ER, EER, ORM, and ORM2, and intended to boost 
interoperability of common conceptual data modelling languages. It was originally designed in UML with textual constraints, and
in this report we present its formalisations in FOL and OWL, which accompanies the paper that describes, discusses, and analyses the KF metamodel in detail.
These new formalizations contribute to give a precise meaning to the metamodel, to understand its complexity properties and to provide a basis for future
implementations.
\end{abstract}

\section{Introduction}

Interoperability of conceptual data modelling languages has become a necessity with ever complex software systems, be this through integration 
of `legacy' systems or {\em de novo} development. In such cases, typically, one would use various conceptual models, such as ORM for usability with 
domain experts and interaction between requirements and data analysis, perhaps an EER model for the back-end database, and UML for any application 
layer software, which requires the system analyst to link entities across models represented in different conceptual modelling languages. 
To aid this endeavour, we have developed an ontology-driven unifying metamodel of UML v2.4.1, ER, EER, ORM and ORM2, whose static, structural 
entities have been presented in \cite{KF13er}, and an extended version also covering constraint in \cite{KF14dke}. 
More specifically, this concerned UML Class diagrams as specified in the UML Superstructure specification v2.4.1 \cite{UMLspec12}, 
typical ORM as described in \cite{ORMisoDraft12,Halpin08}, and both the original ash the encyclopaedic entries for 
ER and EER \cite{Chen76,Song09,Thalheim09}.

This was modelled in UML Class Diagram notation with textual constraints for purpose of facilitating communication. Here we present the FOL
formalisation of that metamodel to ensure precision of meaning and as a first step toward its computational use, including its necessity
for designing efficient algorithms to computationally verify an inter-model assertion is correct (see also \cite{FK14}). 
Secondly, we have formalised it also in OWL 2 \cite{OWL2rec}, for there are many tools that can process OWL files, hence opening up other avenues of 
the use of the metamodel, such as for categorising entities in extant models. 

The remainder of this technical report presents first the FOL formalisation in Section~\ref{sec:fol} 
and subsequently discusses the modelling decisions for the OWL 2 DL version in Section~\ref{sec:owl}. We close in Section~\ref{sec:concl}.

\section{FOL Formalization}
\label{sec:fol}

The FOL formalization is organised along the subfigures of the metamodel, where we first present a brief overview (Section~\ref{sec:overview}), 
and subsequently the relationships and attributes and the constraints among those entities. From Section~\ref{sec:mand} onwards, the model 
constraints are presented, starting with simple mandatory up to ORM's join constraints. Each subfigure of the metamodel is introduced first, 
which is followed by a brief description of the element in the figure and its formalisation. We use  
function-free FOL with equality, 
with a model-theoretic semantics; for a good overview of the language, see, e.g., \cite{Hedman04}.
For notational convenience, we also use counting quantifiers $\exists^{\leq c}$, $\exists^{\geq c}$, and $\exists^{= c}$ for all $c>0$ which do
not add expressive power to FOL but facilitates the interpretation of formula. See 
\cite{Baader08} for their translation into traditional FOL.

\subsection{Overview of the Static Entities}
\label{sec:overview}

\begin{figure}[h]
\centering
   \includegraphics[width=1.0\textwidth]{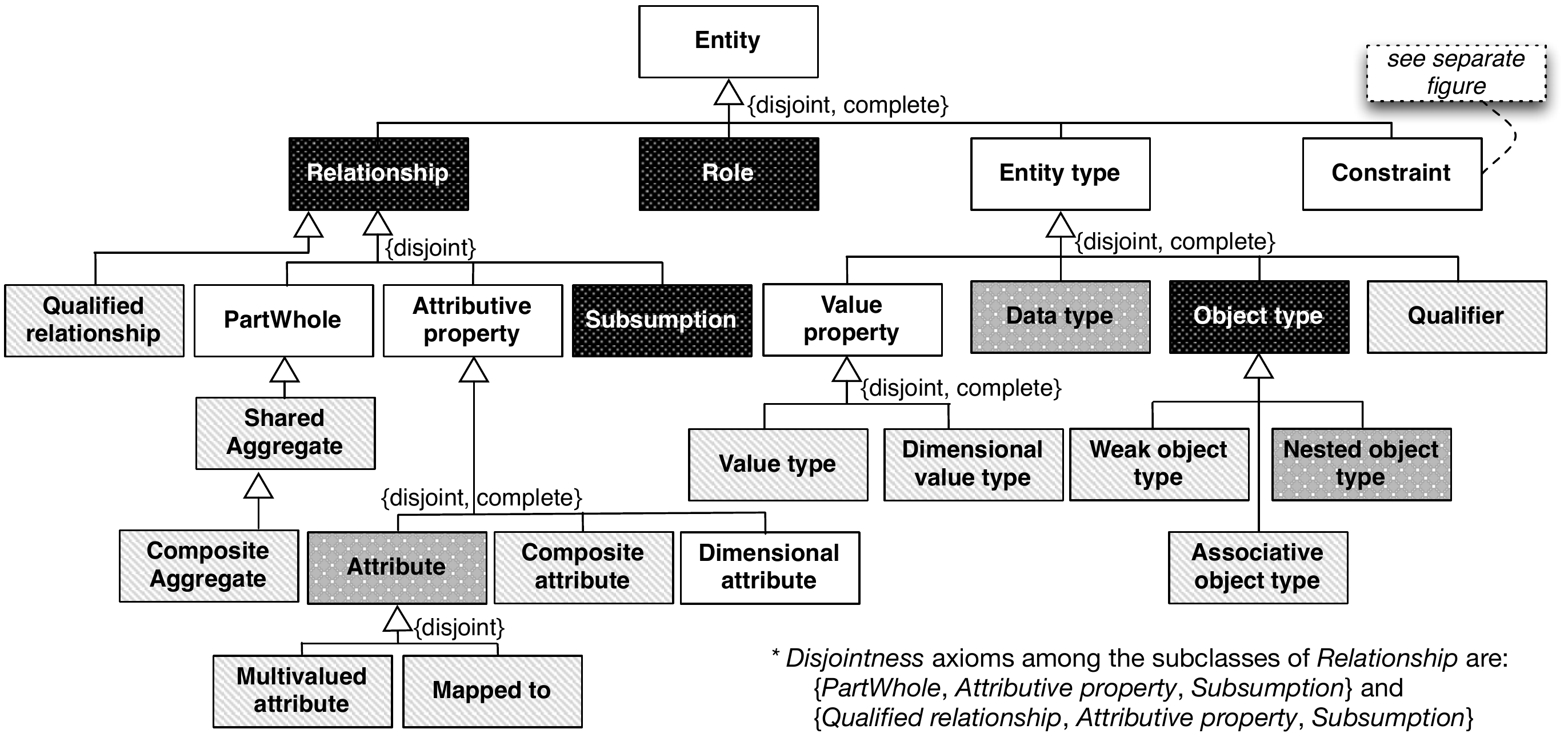} 
    \caption{Principal static entities of the metamodel.}
    \label{fig:entities}
\end{figure}

Figures~\ref{fig:entities} and~\ref{fig:constraints} present the static entities of the metamodel, noting that it has been 
extended cf. \cite{KF13er} with, notably, UML's qualifier, qualified association, and qualified identification. 
As before, we also use these two figures to communicate the overlap among the selected conceptual modelling languages, where a dark colour 
indicates all three families have that entity, classes filled with crossed lines are entities that appear in two of the three, a single line in only one, 
and an non-filled (white) class icon denotes an entity that appears in neither, but is used only in the metamodel to unify the other entities.

\paragraph{Formalization of Static Entities}

The formalization is described based on each UML class diagram element in the corresponding figure, from top to bottom and left to right, and after that
the textual constraints. For example,
in figure \ref{fig:entities} we have 10 isa relations and one textual constraint.

\begin{itemize}
 \item isa (disjoint and complete)

\begin{tabular}{>{\footnotesize}p{13cm}}
$ \forall(x)(\mathtt{Relationship}(x)\rightarrow\mathtt{Entity}(x))$ \\
 $ \forall(x)(\mathtt{Role}(x)\rightarrow\mathtt{Entity}(x)) $ \\
 $ \forall(x)(\mathtt{EntityType}(x)\rightarrow\mathtt{Entity}(x))$  \\
 $ \forall(x)(\mathtt{Constraint}(x)\rightarrow\mathtt{Entity}(x))$\\ 
 $ \forall(x)(\neg(\mathtt{Relationship}(x)\wedge\mathtt{Role}(x)))$ \\
 $ \forall(x)(\neg(\mathtt{Relationship}(x)\wedge\mathtt{EntityType}(x)))$ \\
 $ \forall(x)(\neg(\mathtt{Relationship}(x)\wedge\mathtt{Constraint}(x)))$\\
 $ \forall(x)(\neg(\mathtt{Role}(x)\wedge\mathtt{EntityType}(x))) $\\
 $\forall(x)(\neg(\mathtt{Role}(x)\wedge\mathtt{Constraint}(x)))$\\
 $ \forall(x)(\neg(\mathtt{EntityType}(x)\wedge\mathtt{Constraint}(x)))$\\
 $ \forall(x)(\mathtt{Entity}(x)\rightarrow(\mathtt{Relationship}(x)\vee\mathtt{Role}(x)\vee\mathtt{EntityType}(x)\vee\mathtt{Constraint}(x)))$
\end{tabular}

\item isa 

\begin{tabular}{>{\footnotesize}p{13cm}}
 $ \forall(x)(\mathtt{QualifiedRelationship}(x)\rightarrow\mathtt{Relationship}(x))$
\end{tabular}

 \item isa (disjoint)

\begin{tabular}{>{\footnotesize}p{13cm}}
 $ \forall(x)(\mathtt{PartWhole}(x)\rightarrow\mathtt{Relationship}(x))$\\
 $ \forall(x)(\mathtt{AttributiveProperty}(x)\rightarrow\mathtt{Relationship}(x))$\\
 $ \forall(x)(\mathtt{Subsumption}(x)\rightarrow\mathtt{Relationship}(x))$\\
 $ \forall(x)(\neg(\mathtt{PartWhole}(x)\wedge\mathtt{AttributiveProperty}(x)))$\\
 $ \forall(x)(\neg(\mathtt{PartWhole}(x)\wedge\mathtt{Subsumption}(x)))$\\
 $ \forall(x)(\neg(\mathtt{AttributiveProperty}(x)\wedge\mathtt{Subsumption}(x)))$
\end{tabular}

 \item isa

\begin{tabular}{>{\footnotesize}p{13cm}}
 $ \forall(x)(\mathtt{SharedAggregate}(x)\rightarrow\mathtt{PartWhole}(x))$ 
\end{tabular}

\item isa

\begin{tabular}{>{\footnotesize}p{13cm}}
$ \forall(x)(\mathtt{CompositeAggregate}(x)\rightarrow\mathtt{SharedAggregate}(x))$
\end{tabular}

 \item isa (disjoint, complete)

\begin{tabular}{>{\footnotesize}p{13cm}}
$\forall(x)(\mathtt{Attribute}(x)\rightarrow\mathtt{AttributiveProperty}(x))$\\
$\forall(x)(\mathtt{CompositeAttribute}(x)\rightarrow\mathtt{AttributiveProperty}(x))$\\
  $ \forall(x)(\mathtt{DimensionalAttribute}(x)\rightarrow\mathtt{AttributiveProperty}(x))$\\
  $ \forall(x)(\neg(\mathtt{Attribute}(x)\wedge\mathtt{CompositeAttribute}(x)))$\\
  $ \forall(x)(\neg(\mathtt{Attribute}(x)\wedge\mathtt{DimensionalAttribute}(x)))$\\
  $ \forall(x)(\neg(\mathtt{CompositeAttribute}(x)\wedge\mathtt{DimensionalAttribute}(x)))$\\
 $\forall(x)(\mathtt{AttributiveProperty}(x)\rightarrow(\mathtt{Attribute}(x)\vee\mathtt{CompositeAttribute}(x)\vee$\\
 \hspace{1.8cm}$\mathtt{DimensionalAttribute}(x)))$             
\end{tabular}
  
\item isa (disjoint)

\begin{tabular}{>{\footnotesize}p{13cm}}
 $ \forall(x)(\mathtt{MultivaluedAttribute}(x)\rightarrow\mathtt{Attribute}(x))$\\
 $ \forall(x)(\mathtt{MappedTo}(x)\rightarrow\mathtt{Attribute}(x))$\\
 $ \forall(x)(\neg(\mathtt{MultivaluedAttribute}(x)\wedge\mathtt{MappedTo}(x)))$
\end{tabular}

\item isa (disjoint, complete)

\begin{tabular}{>{\footnotesize}p{13cm}}
 $ \forall(x)(\mathtt{ValueProperty}(x)\rightarrow\mathtt{EntityType}(x))$\\
 $ \forall(x)(\mathtt{DataType}(x)\rightarrow\mathtt{EntityType}(x))$\\
 $ \forall(x)(\mathtt{ObjectType}(x)\rightarrow\mathtt{EntityType}(x))$\\
 $ \forall(x)(\mathtt{Qualifier}(x)\rightarrow\mathtt{EntityType}(x))$\\
 $ \forall(x)(\neg(\mathtt{ValueProperty}(x)\wedge\mathtt{DataType}(x)))$\\
 $ \forall(x)(\neg(\mathtt{ValueProperty}(x)\wedge\mathtt{ObjectType}(x)))$\\
 $ \forall(x)(\neg(\mathtt{ValueProperty}(x)\wedge\mathtt{Qualifier}(x)))$\\
 $ \forall(x)(\neg(\mathtt{DataType}(x)\wedge\mathtt{ObjectType}(x)))$\\
 $ \forall(x)(\neg(\mathtt{DataType}(x)\wedge\mathtt{Qualifier}(x)))$\\
 $ \forall(x)(\neg(\mathtt{ObjectType}(x)\wedge\mathtt{Qualifier}(x)))$\\
 $ \forall(x)(\mathtt{EntityType}(x)\rightarrow(\mathtt{ValueProperty}(x)\vee\mathtt{DataType}(x)\vee\mathtt{ObjectType}(x)\vee
 \mathtt{Qualifier}(x)))$
\end{tabular}
 
\item isa (disjoint,complete)

\begin{tabular}{>{\footnotesize}p{13cm}}
 $ \forall(x)(\mathtt{ValueType}(x)\rightarrow\mathtt{ValueProperty}(x))$\\
 $ \forall(x)(\mathtt{DimensionalValueType}(x)\rightarrow\mathtt{ValueProperty}(x))$\\
 $ \forall(x)(\neg(\mathtt{ValueType}(x)\wedge\mathtt{DimensionalValueType}(x)))$\\
 $ \forall(x)(\mathtt{ValueProperty}(x)\rightarrow(\mathtt{ValueType}(x)\vee\mathtt{DimensionalValueType}(x)))$
\end{tabular}

\item isa

\begin{tabular}{>{\footnotesize}p{13cm}}
 $ \forall(x)(\mathtt{WeakObjectType}(x)\rightarrow\mathtt{ObjectType}(x))$\\
 $ \forall(x)(\mathtt{NestedObjectType}(x)\rightarrow\mathtt{ObjectType}(x))$\\
 $ \forall(x)(\mathtt{AssociativeObjectType}(x)\rightarrow\mathtt{ObjectType}(x))$
\end{tabular}

\item disjointness (first textual constraint)

\begin{tabular}{>{\footnotesize}p{13cm}}
 $ \forall(x)(\neg(\mathtt{QualifiedRelationship}(x)\wedge\mathtt{AttributiveProperty}(x)))$\\
 $ \forall(x)(\neg(\mathtt{QualifiedRelationship}(x)\wedge\mathtt{Subsumption}(x)))$ \\
 $ \forall(x)(\neg(\mathtt{AttributiveProperty}(x)\wedge\mathtt{Subsumption}(x)))$
\end{tabular}

\end{itemize}

\subsection{Overview of the Constraints}

\begin{figure}[h]
\centering
   \includegraphics[width=1.0\textwidth]{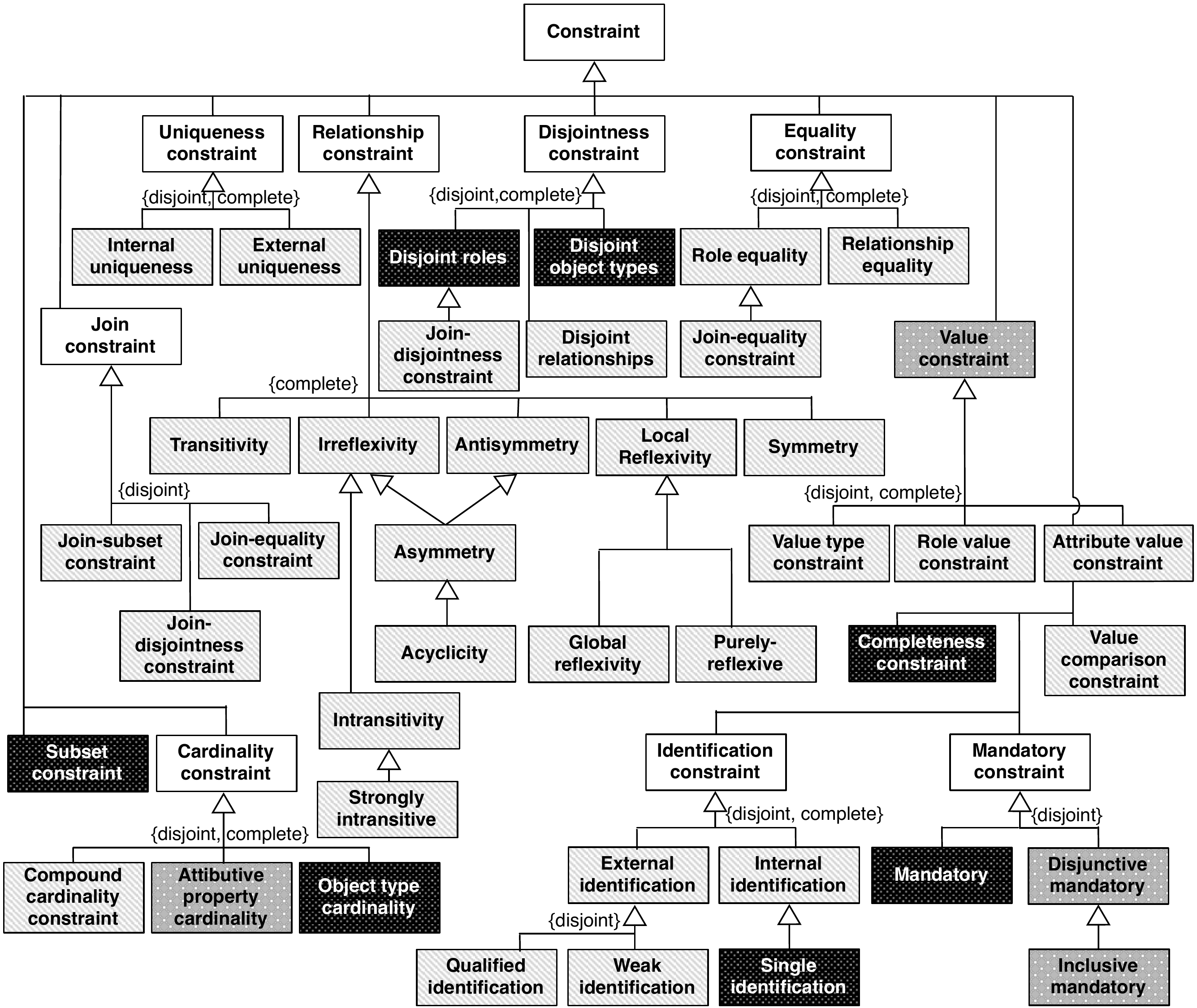}
    \caption{Unified hierarchy of constraints in the metamodel.} 
    \label{fig:constraints}
\end{figure}

There are several types of constraints in modelling languages, and this is reflected in the number of nodes in the graph in figure \ref{fig:constraints}. The main
is relation in the figue does not show all the siblings at the same level due to space limitations. In this section we only show the different types of constraints.
In later sections, each constraint type is analyzed in detail together with related static entities.

\paragraph{Formalization of Constraint types}

\begin{itemize}
\item isa

\begin{tabular}{>{\footnotesize}p{13cm}}
  $ \forall(x)(\mathtt{CardinalityConstraint}(x)\rightarrow\mathtt{Constraint}(x))$\\
  $ \forall(x)(\mathtt{SubsetConstraint}(x)\rightarrow\mathtt{Constraint}(x))$\\
  $ \forall(x)(\mathtt{JoinConstraint}(x)\rightarrow\mathtt{Constraint}(x))$\\
  $ \forall(x)(\mathtt{UniquenessConstraint}(x)\rightarrow\mathtt{Constraint}(x))$\\
  $ \forall(x)(\mathtt{RelationshipConstraint}(x)\rightarrow\mathtt{Constraint}(x))$\\
  $ \forall(x)(\mathtt{DisjointnessConstraint}(x)\rightarrow\mathtt{Constraint}(x))$\\
  $ \forall(x)(\mathtt{EqualityConstraint}(x)\rightarrow\mathtt{Constraint}(x))$\\
  $ \forall(x)(\mathtt{ValueConstraint}(x)\rightarrow\mathtt{Constraint}(x))$\\
  $ \forall(x)(\mathtt{CompletenessConstraint}(x)\rightarrow\mathtt{Constraint}(x))$\\
  $ \forall(x)(\mathtt{ValueComparisonConstraint}(x)\rightarrow\mathtt{Constraint}(x))$\\
  $ \forall(x)(\mathtt{IdentificationConstraint}(x)\rightarrow\mathtt{Constraint}(x))$\\
  $ \forall(x)(\mathtt{MandatoryConstraint}(x)\rightarrow\mathtt{Constraint}(x))$
\end{tabular}

\item isa (disjoint, complete)

\begin{tabular}{>{\footnotesize}p{13cm}}
  $ \forall(x)(\mathtt{CompoundCardinalityConstraint}(x)\rightarrow\mathtt{CardinalityConstraint}(x))$\\
  $ \forall(x)(\mathtt{AttributivePropertyCardinality}(x)\rightarrow\mathtt{CardinalityConstraint}(x))$\\
  $ \forall(x)(\mathtt{ObjectTypeCardinality}(x)\rightarrow\mathtt{CardinalityConstraint}(x))$\\
   $ \forall(x)(\neg(\mathtt{CompoundCardinalityConstraint}(x)\wedge 
   \mathtt{AttributivePropertyCardinality}(x)))$\\
    $ \forall(x)(\neg(\mathtt{CompoundCardinalityConstraint}(x)\wedge\mathtt{ObjectTypeCardinality}(x)))$\\
     $ \forall(x)(\neg(\mathtt{AttributivePropertyCardinality}(x)\wedge\mathtt{ObjectTypeCardinality}(x)))$\\
 $ \forall(x)(\mathtt{CardinalityConstraint}(x)\rightarrow(\mathtt{CompoundCardinalityConstraint}(x)\vee$\\
 \hspace{1.8cm}$\mathtt{AttributivePropertyCardinality}(x)\vee\mathtt{ObjectTypeCardinality}(x)))$
\end{tabular}

 \item isa (disjoint)

\begin{tabular}{>{\footnotesize}p{13cm}}
   $ \forall(x)(\mathtt{JoinSubsetConstraint}(x)\rightarrow\mathtt{JoinConstraint}(x))$\\
$ \forall(x)(\mathtt{JoinEqualityConstraint}(x)\rightarrow\mathtt{JoinConstraint}(x))$\\
    $ \forall(x)(\mathtt{JoinDisjointnessConstraint}(x)\rightarrow\mathtt{JoinConstraint}(x))$\\
    $ \forall(x)(\neg(\mathtt{JoinSubsetConstraint}(x)\wedge\mathtt{JoinEqualityConstraint}(x)))$\\
    $ \forall(x)(\neg(\mathtt{JoinSubsetConstraint}(x)\wedge\mathtt{JoinDisjointnessConstraint}(x)))$\\
        $ \forall(x)(\neg(\mathtt{JoinEqualityConstraint}(x)\wedge\mathtt{JoinDisjointnessConstraint}(x)))$  
\end{tabular}
        
 \item isa (disjoint, complete)

\begin{tabular}{>{\footnotesize}p{13cm}}
  $ \forall(x)(\mathtt{InternalUniquenessConstraint}(x)\rightarrow\mathtt{UniquenessConstraint}(x))$\\
    $ \forall(x)(\mathtt{ExternalUniquenessConstraint}(x)\rightarrow\mathtt{UniquenessConstraint}(x))$\\
    $ \forall(x)(\neg(\mathtt{InternalUniquenessConstraint}(x)\wedge 
    \mathtt{ExternalUniquenessConstraint}(x)))$\\
 $ \forall(x)(\mathtt{UniquenessConstraint}(x)\rightarrow(\mathtt{InternalUniquenessConstraint}(x)\vee$\\
 \hspace{1.8cm}$\mathtt{ExternalUniquenessConstraint}(x)))$
\end{tabular}

 \item isa (complete)

\begin{tabular}{>{\footnotesize}p{13cm}}
   $ \forall(x)(\mathtt{Transitivity}(x)\rightarrow\mathtt{RelationshipConstraint}(x))$\\
    $ \forall(x)(\mathtt{Antisymmetry}(x)\rightarrow\mathtt{RelationshipConstraint}(x))$\\
  $ \forall(x)(\mathtt{Irreflexivity}(x)\rightarrow\mathtt{RelationshipConstraint}(x))$\\
  $ \forall(x)(\mathtt{LocalReflexivity}(x)\rightarrow\mathtt{RelationshipConstraint}(x))$\\
  $ \forall(x)(\mathtt{Symmetry}(x)\rightarrow\mathtt{RelationshipConstraint}(x))$\\
 $ \forall(x)(\mathtt{RelationshipConstraint}(x)\rightarrow(\mathtt{Transitivity}(x)\vee\mathtt{Antisymmetry}(x)\vee$\\
 \hspace{1.8cm}$\mathtt{Irreflexivity}(x)\vee\mathtt{LocalReflexivity}\vee\mathtt{Symmetry}(x)))$
\end{tabular}

\item isa

\begin{tabular}{>{\footnotesize}p{13cm}}
  $ \forall(x)(\mathtt{Asymmetry}(x)\rightarrow\mathtt{Antisymmetry}(x))$
\end{tabular}

\item isa

\begin{tabular}{>{\footnotesize}p{13cm}}
  $ \forall(x)(\mathtt{Asymmetry}(x)\rightarrow\mathtt{Irreflexivity}(x))$
\end{tabular}

\item isa

\begin{tabular}{>{\footnotesize}p{13cm}}
  $ \forall(x)(\mathtt{Acyclicity}(x)\rightarrow\mathtt{Asymmetry}(x))$
\end{tabular}

\item isa

\begin{tabular}{>{\footnotesize}p{13cm}}
  $ \forall(x)(\mathtt{Intransitivity}(x)\rightarrow\mathtt{Irreflexivity}(x))$
\end{tabular}

\item isa

\begin{tabular}{>{\footnotesize}p{13cm}}
  $ \forall(x)(\mathtt{StronglyIntransitivity}(x)\rightarrow\mathtt{Intransitivity}(x))$
\end{tabular}

\item isa

\begin{tabular}{>{\footnotesize}p{13cm}}
  $ \forall(x)(\mathtt{GlobalReflexivity}(x)\rightarrow\mathtt{LocalReflexivity}(x))$
\end{tabular}

\item isa

\begin{tabular}{>{\footnotesize}p{13cm}}
  $ \forall(x)(\mathtt{PurelyReflexivity}(x)\rightarrow\mathtt{LocalReflexivity}(x))$
\end{tabular}

\item isa (disjoint, complete)

\begin{tabular}{>{\footnotesize}p{13cm}}
 $ \forall(x)(\mathtt{DisjointRoles}(x)\rightarrow\mathtt{DisjointnessConstraint}(x))$\\
  $ \forall(x)(\mathtt{DisjointRelationships}(x)\rightarrow\mathtt{DisjointnessConstraint}(x))$\\ 
     $ \forall(x)(\mathtt{DisjointObjectTypes}(x)\rightarrow\mathtt{DisjointnessConstraint}(x))$\\
  $ \forall(x)(\neg(\mathtt{DisjointRoles}(x)\wedge\mathtt{DisjointObjectTypes}(x)))$\\
  $ \forall(x)(\neg(\mathtt{DisjointRoles}(x)\wedge\mathtt{DisjointRelationships}(x)))$\\
    $ \forall(x)(\neg(\mathtt{DisjointRelationships}(x)\wedge\mathtt{DisjointObjectTypes}(x)))$\\  
 $ \forall(x)(\mathtt{DisjointnessConstraint}(x)\rightarrow(\mathtt{DisjointRoles}(x)\vee$\\
 \hspace{1.8cm}$\mathtt{DisjointObjectTypes}(x)\vee\mathtt{DisjointRelationships}(x)))$
\end{tabular}

\item isa

\begin{tabular}{>{\footnotesize}p{13cm}}
  $ \forall(x)(\mathtt{JoinDisjointnessConstraint}(x)\rightarrow\mathtt{DisjointRoles}(x))$ 
\end{tabular}

\item isa (disjoint, complete)

\begin{tabular}{>{\footnotesize}p{13cm}}
$ \forall(x)(\mathtt{RoleEquality}(x)\rightarrow\mathtt{EqualityConstraint}(x))$\\
       $ \forall(x)(\mathtt{RelationshipEquality}(x)\rightarrow\mathtt{EqualityConstraint}(x))$\\
 $ \forall(x)(\neg(\mathtt{RoleEquality}(x)\wedge\mathtt{RelationshipEquality}(x)))$\\
 $ \forall(x)(\mathtt{EqualityConstraint}(x)\rightarrow(\mathtt{RoleEquality}(x)\vee 
 \mathtt{RelationshipEquality}(x)))$          
\end{tabular}

\item isa

\begin{tabular}{>{\footnotesize}p{13cm}}
$ \forall(x)(\mathtt{JoinEqualityConstraint}(x)\rightarrow\mathtt{RoleEquality}(x))$ 
\end{tabular}

\item isa (disjoint, complete)

\begin{tabular}{>{\footnotesize}p{13cm}}
$ \forall(x)(\mathtt{ValueTypeConstraint}(x)\rightarrow\mathtt{ValueConstraint}(x))$\\
    $ \forall(x)(\mathtt{RoleValueConstraint}(x)\rightarrow\mathtt{ValueConstraint}(x))$\\
    $ \forall(x)(\mathtt{AttributeValueConstraint}(x)\rightarrow\mathtt{ValueConstraint}(x))$\\
 $ \forall(x)(\neg(\mathtt{ValueTypeConstraint}(x)\wedge\mathtt{RoleValueConstraint}(x)))$\\
  $ \forall(x)(\neg(\mathtt{ValueTypeConstraint}(x)\wedge\mathtt{AttributeValueConstraint}(x)))$\\
   $ \forall(x)(\neg(\mathtt{RoleValueConstraint}(x)\wedge\mathtt{AttributeValueConstraint}(x)))$\\
 $ \forall(x)(\mathtt{ValueConstraint}(x)\rightarrow(\mathtt{ValueTypeConstraint}(x)\vee$\\
 \hspace{1.8cm}$\mathtt{RoleValueConstraint}(x)\vee\mathtt{AttributeValueConstraint}(x)))$          
\end{tabular}

\item isa (disjoint, complete)

\begin{tabular}{>{\footnotesize}p{13cm}}
    $ \forall(x)(\mathtt{ExternalIdentification}(x)\rightarrow\mathtt{IdentificationConstraint}(x))$\\
    $ \forall(x)(\mathtt{InternalIdentification}(x)\rightarrow\mathtt{IdentificationConstraint}(x))$\\
 $ \forall(x)(\neg(\mathtt{ExternalIdentification}(x)\wedge\mathtt{InternalIdentification}(x)))$\\
 $ \forall(x)(\mathtt{IdentificationConstraint}(x)\rightarrow(\mathtt{ExternalIdentification}(x)\vee$\\
 \hspace{1.8cm}$\mathtt{InternalIdentification}(x)))$
\end{tabular}

\item isa (disjoint)

\begin{tabular}{>{\footnotesize}p{13cm}}
$ \forall(x)(\mathtt{QualifiedIdentification}(x)\rightarrow\mathtt{ExternalIdentification}(x))$\\
        $ \forall(x)(\mathtt{WeakIdentification}(x)\rightarrow\mathtt{ExternalIdentification}(x))$\\
 $ \forall(x)(\neg(\mathtt{QualifiedIdentification}(x)\wedge\mathtt{WeakIdentification}(x)))$
\end{tabular}

\item isa

\begin{tabular}{>{\footnotesize}p{13cm}}
$ \forall(x)(\mathtt{SingleIdentification}(x)\rightarrow\mathtt{InternalIdentification}(x))$ 
\end{tabular}

\item isa (disjoint)

\begin{tabular}{>{\footnotesize}p{13cm}}
$\forall(x)(\mathtt{Mandatory}(x)\rightarrow\mathtt{MandatoryConstraint}(x))$\\
     $ \forall(x)(\mathtt{DisjunctiveMandatory}(x)\rightarrow\mathtt{MandatoryConstraint}(x))$\\
     $\forall(x)(\neg(\mathtt{MandatoryConstraint}(x)\wedge\mathtt{DisjunctiveMandatory}(x)))$
\end{tabular}

\item isa
       
\begin{tabular}{>{\footnotesize}p{13cm}}
$\forall(x)(\mathtt{InclusiveMandatory}(x)\rightarrow\mathtt{DisjunctiveMandatory}(x))$
\end{tabular}

\end{itemize}

\subsection{Relationships between Relationship, Role and Entity type}

\begin{figure}[t]
\centering
   \includegraphics[width=0.75\textwidth]{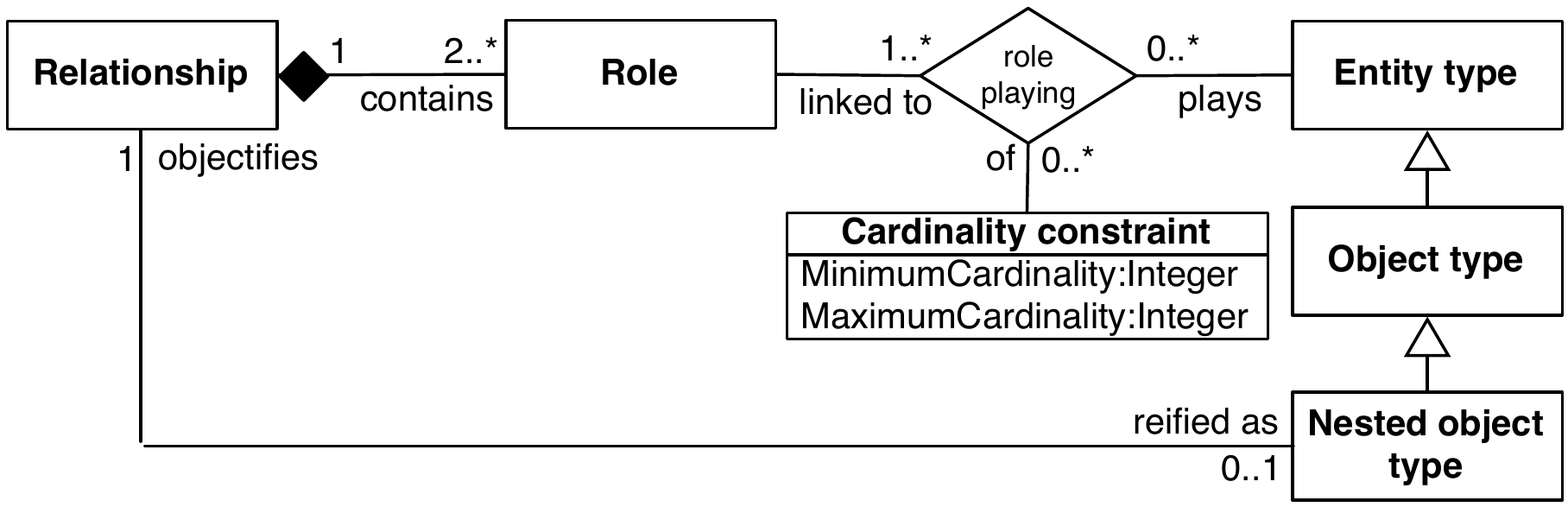}
    \caption{Relationships between {\sf Relationship}, {\sf Role}, and {\sf Entity type}; see text for details.}
    \label{fig:roleRel}
\end{figure}

This section goes into detail on relationships and roles in general, and subsequently subsumption and aggregation, and attributes and value types.
The main entities and constraints pertaining to relationships are depicted in Figure~\ref{fig:roleRel}.

\paragraph{Formalization of Relationship, Role, EntityType and related constraints}

\begin{itemize}
\item composition $(1,2..*)$

\begin{tabular}{>{\footnotesize}p{13cm}}
 $ \forall(x,y)(\mathtt{Contains}(x,y)\rightarrow\mathtt{Relationship}(x)\wedge\mathtt{Role}(y))$\\
 $ \forall(x)\exists^{\geq 2}y(\mathtt{Contains}(x,y))$\\
 $ \forall(x)(\mathtt{Role}(x)\rightarrow\exists^{=1}(y)(\mathtt{Contains}(y,x)))$
\end{tabular}

\noindent
We do not need to add asymmetric and irreflexive axioms to the formalization because they are both implicit after the disjointness between $\mathtt{Relationship}$ and $\mathtt{Role}$.
Also, transitivity is not included since we are dealing with a ``direct parthood'' relationship.

\item ternary relation $(1..*,0..*,0..*)$

\begin{tabular}{>{\footnotesize}p{13cm}}
$ \forall(x,y,z)(\mathtt{RolePlaying}(x,y,z)\rightarrow(\mathtt{Role}(x)\wedge\mathtt{CardinalityConstraint}(y)\wedge
\mathtt{EntityType}(z)))$\\
 $ \forall(x)(\mathtt{Role}(x)\rightarrow\exists^{\geq 1}(y,z)(\mathtt{RolePlaying}(x,y,z)))$
\end{tabular}

\noindent
There are several possible interpretations of cardinality constraints over non-binary relations on UML class diagrams. In this formalization we interpret
all such constraints as in \cite{Artale07er}, constraining the quantity of tuples in which each entity may participate. This choice is reflected in the formula.

\item attributes

\begin{tabular}{>{\footnotesize}p{13cm}}
 $\forall(x,y)((\mathtt{CardinalityConstraint}(x)\wedge\mathtt{MinimumCardinality}(x,y))\rightarrow\mathtt{Integer}(y))$\\
 $\forall(x,y)((\mathtt{CardinalityConstraint}(x)\wedge\mathtt{MaximumCardinality}(x,y))\rightarrow\mathtt{Integer}(y))$\\
 $\forall(x)(\mathtt{CardinalityConstraint}(x)\rightarrow\exists^{\leq 1}(y)(\mathtt{MinimumCardinality}(x,y)))$\\
 $\forall(x)(\mathtt{CardinalityConstraint}(x)\rightarrow\exists^{\leq 1}(y)(\mathtt{MaximumCardinality}(x,y)))$ 
\end{tabular}

\noindent
Attributes in UML classes may also have cardinality constraints. In this figure, we understand each attribute to 
have $[0..1]$ cardinality. The type $\mathtt{Integer}$ is assumed to be a range of integer numbers from $0$ to a maximum number noted $n$. This last
number is interpreted to mean ``any integer value''.

\item relation $(1,0..1)$

\begin{tabular}{>{\footnotesize}p{13cm}}
$ \forall(x,y)(\mathtt{ReifiedAs}(x,y)\rightarrow\mathtt{Relationship}(x)\wedge\mathtt{NestedObjectType}(y))$\\
 $ \forall(x)(\mathtt{NestedObjectType}(x)\rightarrow\exists^{=1}(y)(\mathtt{ReifiedAs}(x,y)))$\\
 $ \forall(x(\mathtt{Relationship}(x)\rightarrow\exists^{\leq 1}(y)(\mathtt{ReifiedAs}(x,y)))$
\end{tabular}

\item relation between ReifiedAs, Contains and RolePlaying (not in figure)

\begin{tabular}{>{\footnotesize}p{13cm}}
$ \forall(x,y)(\mathtt{ReifiedAs}(x,y)\rightarrow\forall(z,w)(\mathtt{Contains}(x,z)\leftrightarrow\mathtt{RolePlaying}(z,w,y)))^{*}$
\end{tabular}

\noindent
This constraint has more than two variables and cannot be replaced with equivalent formula in two variables. All the formula with this
property are shown with a $^*$ at then end. This star means the formula has to be taken out the formalization in
order to show its decidability. See section \ref{sec:properties} for more on the properties of the formalization.
\end{itemize}

\subsubsection{Subsumption and Aggregation}

\begin{figure}[t]
\centering
   \includegraphics[width=0.95\textwidth]{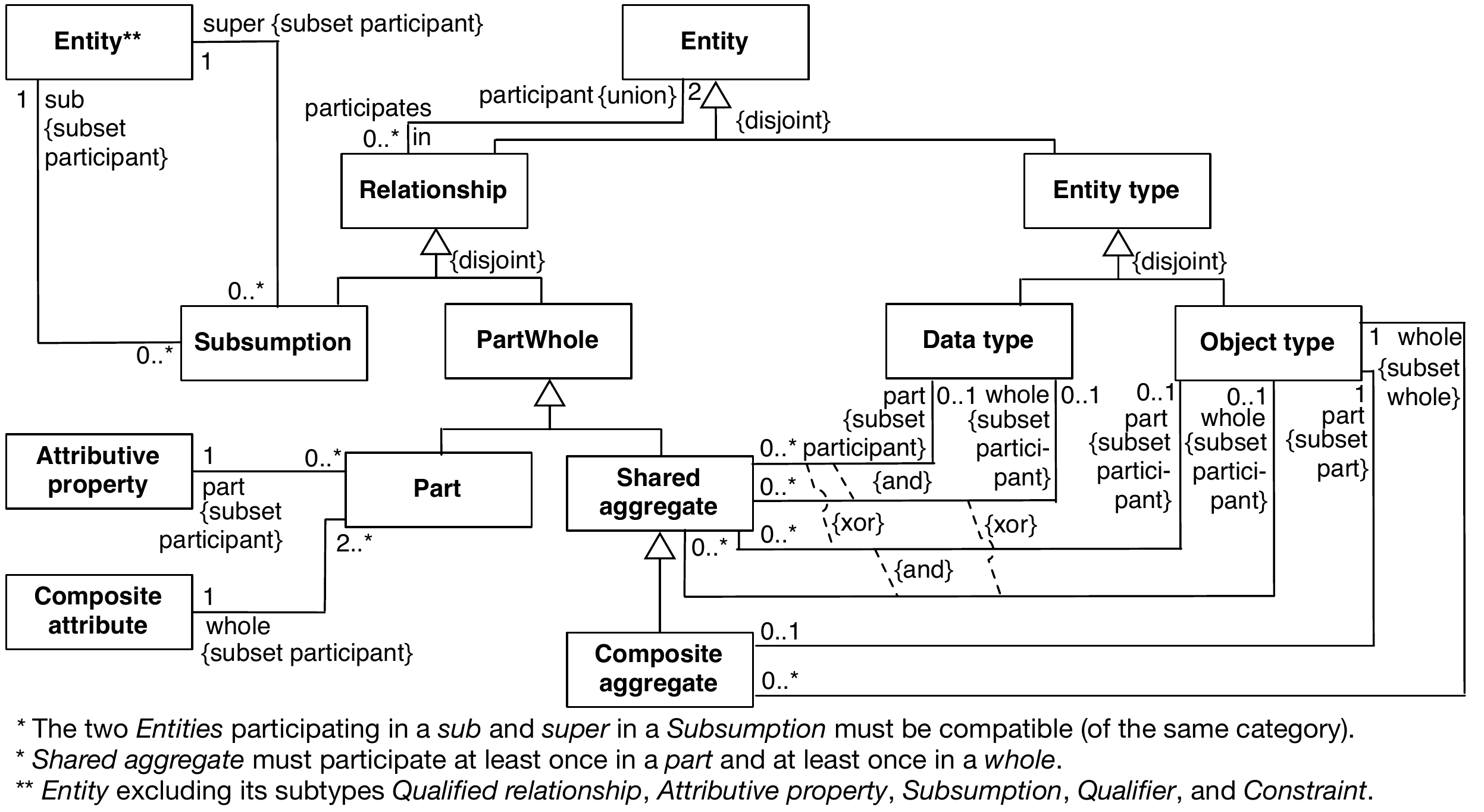}
    \caption{Subsumption and aggregation.} 
    \label{fig:isa}
\end{figure}

The metamodel fragment dealing with subsumption and aggregation is depicted in Figure~\ref{fig:isa} and has the following logic-based reconstruction.

\paragraph{Formalization of Subsumption and Aggregation}

\begin{itemize}
\item relation $(0..*,2)$, subset  and union constraints

\begin{tabular}{>{\footnotesize}p{13cm}}
 $\forall(x,y)(\mathtt{Participant}(x,y)\rightarrow((\mathtt{Relationship}(x)\wedge\mathtt{Entity}(y))\vee$\\
 \hspace{1.8cm}$(\mathtt{ValueComparisonConstraint}(x)\wedge\mathtt{Role}(y))\vee$\\
\hspace{1.8cm}$(\mathtt{RelationshipConstraint}(x)\wedge\mathtt{Role}(y))))$\\
 $\forall(x)(\mathtt{Relationship}(x)\rightarrow\exists^{=2}(y)(\mathtt{Participant}(x,y)))$\\
 $\forall(x,y)((\mathtt{Participant}(x,y)\wedge\mathtt{Relationship}(x))\leftrightarrow(\mathtt{Sub}(x,y)\vee$\\
\hspace{1.8cm}$\mathtt{Super}(x,y)\vee\mathtt{Part}(x,y)\vee\mathtt{Whole}(x,y)))$
\end{tabular}

\item relation $(1,0..*)$,  third textual constraint

\begin{tabular}{>{\footnotesize}p{13cm}}
  $ \forall(x,y)(\mathtt{Sub}(x,y)\rightarrow(\mathtt{Subsumption}(x)\wedge\mathtt{Entity}(y)\wedge$\\
 \hspace{1.8cm}$\neg\mathtt{QualifiedRelationship}(y)\wedge\neg\mathtt{AttributiveProperty}(y)\wedge$\\
 \hspace{1.8cm}$\neg\mathtt{Subsumption}(y)\wedge\neg\mathtt{Qualifier}(y)\wedge\neg\mathtt{Constraint}(y)))$\\
  $ \forall(x)(\mathtt{Subsumption}(x)\rightarrow\exists^{=1}(y)(\mathtt{Sub}(x,y)))$
 \end{tabular}

  \item relation $(1,0..*)$, third textual constraint

\begin{tabular}{>{\footnotesize}p{13cm}}
 $ \forall(x,y)(\mathtt{Super}(x,y)\rightarrow(\mathtt{Subsumption}(x)\wedge\mathtt{Entity}(y)\wedge$\\
\hspace{1.8cm}$\neg\mathtt{QualifiedRelationship}(y)\wedge\neg\mathtt{AttributiveProperty}(y)\wedge$\\
\hspace{1.8cm}$\neg\mathtt{Subsumption}(y)\wedge\neg\mathtt{Qualifier}(y)\wedge\neg\mathtt{Constraint}(y)))$\\ 
  $ \forall(x)(\mathtt{Subsumption}(x)\rightarrow\exists^{=1}(y)(\mathtt{Super}(x,y)))$
\end{tabular}

 \item disjunctive exclusion (not in figure)

\begin{tabular}{>{\footnotesize}p{13cm}}
 $\forall(x,y)((\mathtt{Subsumption}(x)\wedge\mathtt{Sub}(x,y)\wedge\Phi(y))\rightarrow\exists(z)(\Phi(z)\wedge\mathtt{Super}(x,z)))^*$\\
 $\forall(x,y)((\mathtt{Subsumption}(x)\wedge\mathtt{Super}(x,y)\wedge\Phi(y))\rightarrow\exists(z)(\Phi(z)\wedge\mathtt{Sub}(x,z)))^*$
\end{tabular}

\noindent
For all $\Phi \in \{\mathtt{SharedAggregate}, \mathtt{CompositeAggregate}, \mathtt{Role}, \mathtt{DataType}, $\\
$\mathtt{Relationship}\wedge\neg\mathtt{PartWhole},\mathtt{ObjectType}\}$.

\item isa

\begin{tabular}{>{\footnotesize}p{13cm}}
 $\forall(x,y)(\mathtt{Part}(x)\rightarrow\mathtt{PartWhole}(x))$
\end{tabular}

\item relation $(0..*,1)$

\begin{tabular}{>{\footnotesize}p{13cm}}
 $\forall(x,y)(\mathtt{Part}(x,y)\rightarrow((\mathtt{Part}(x)\wedge\mathtt{AttributiveProperty}(y))\vee$\\
 \hspace{1.8cm}$(\mathtt{SharedAggregate}(x)\wedge\mathtt{DataType}(y))\vee$\\
 \hspace{1.8cm}$(\mathtt{SharedAggregate}(x)\wedge\mathtt{ObjectType}(y))))$\\
 $\forall(x)(\mathtt{Part}(x)\rightarrow\exists^{=1}(y)(\mathtt{Part}(x,y)\wedge\mathtt{AttributiveProperty}(y)))$
\end{tabular}

\item relation $(2..*,1)$

\begin{tabular}{>{\footnotesize}p{13cm}}
 $\forall(x,y)(\mathtt{Whole}(x,y)\rightarrow((\mathtt{Part}(x)\wedge\mathtt{CompositeAttribute}(y))\vee$\\
 \hspace{1.8cm}$(\mathtt{SharedAggregate}(x)\wedge\mathtt{DataType}(y))\vee$\\
 \hspace{1.8cm}$(\mathtt{SharedAggregate}(x)\wedge\mathtt{ObjectType}(y))))$\\
 $\forall(x)(\mathtt{Part}(x)\rightarrow\exists^{=1}(y)(\mathtt{Whole}(x,y)\wedge\mathtt{CompositeAttribute}(y)))$\\
 $\forall(x)(\mathtt{CompositeAttribute}(x)\rightarrow\exists^{\geq 2}(y)(\mathtt{Part}(y)\wedge\mathtt{Whole}(y,x)))$
\end{tabular}

\item relation $(0..*,0..1)$, xor, second textual constraint

\begin{tabular}{>{\footnotesize}p{13cm}}
 $ \forall(x)(\mathtt{SharedAggregate}(x)\rightarrow\exists^{\leq 1}(y)((\mathtt{DataType}(y)\vee\mathtt{ObjectType}(y))\wedge
 \mathtt{Part}(x,y)))$\\
 $\forall(x)(\mathtt{SharedAggregate}(x)\rightarrow(\exists^{\geq 1}(y)(\mathtt{Part}(x,y))\wedge\exists^{\geq 1}(y)(\mathtt{Whole}(x,y))))$
\end{tabular}

\item relation $(0..*,0..1)$, xor

\begin{tabular}{>{\footnotesize}p{13cm}}
 $ \forall(x)(\mathtt{SharedAggregate}(x)\rightarrow\exists^{\leq 1}(y)((\mathtt{DataType}(y)\vee\mathtt{ObjectType}(y))\wedge\mathtt{Whole}(x,y)))$\\
\end{tabular}

\noindent 
The last two set of formula only show the disjunction of the xor. The exclusion between $\mathtt{DataType}$ and $\mathtt{ObjectType}$ is inherited from the disjoint isa in figure \ref{fig:entities}.

 \item and relationships

\begin{tabular}{>{\footnotesize}p{13cm}}
$\forall(x,y)((\mathtt{SharedAggregate}(x)\wedge\mathtt{DataType}(y)\wedge\mathtt{Part}(x,y))\rightarrow$\\
\hspace{1.8cm}$\exists(z)(\mathtt{DataType}(z)\wedge\mathtt{Whole}(x,z)))^*$\\
$\forall(x,y)((\mathtt{SharedAggregate}(x)\wedge\mathtt{DataType}(y)\wedge\mathtt{Whole}(x,y))\rightarrow$\\
\hspace{1.8cm}$\exists(z)(\mathtt{DataType}(z)\wedge\mathtt{Part}(x,z)))^*$\\
$\forall(x,y)((\mathtt{SharedAggregate}(x)\wedge\mathtt{ObjectType}(y)\wedge\mathtt{Part}(x,y))\rightarrow$\\
\hspace{1.8cm}$\exists(z)(\mathtt{ObjectType}(z)\wedge\mathtt{Whole}(x,z)))^*$\\
$\forall(x,y)((\mathtt{SharedAggregate}(x)\wedge\mathtt{ObjectType}(y)\wedge\mathtt{Whole}(x,y))\rightarrow$\\
\hspace{1.8cm}$\exists(z)(\mathtt{ObjectType}(z)\wedge\mathtt{Part}(x,z)))^*$
\end{tabular}

\item irreflexivity, asymmetry (not in figure)

\begin{tabular}{>{\footnotesize}p{13cm}}
 $\forall(x,y,z)((\mathtt{PartWhole}(x)\wedge\mathtt{Part}(x,y)\wedge\mathtt{Whole}(x,z))\rightarrow\neg(y=z))^*$\\
 $\forall(x,y,z)((\mathtt{PartWhole}(x)\wedge\mathtt{Part}(x,y)\wedge\mathtt{Whole}(x,z))\rightarrow$\\
\hspace{1.8cm}$\neg\exists(v)(\mathtt{PartWhole}(v)\wedge\mathtt{Part}(v,z)\wedge\mathtt{Whole}(v,y)))^*$
\end{tabular}

\item relation $(0..1,1)$

\begin{tabular}{>{\footnotesize}p{13cm}}
 $\forall(x,y)((\mathtt{Part}(x,y)\wedge\mathtt{CompositeAggregate}(x))\rightarrow\mathtt{ObjectType}(y))$\\
  $\forall(x)(\mathtt{CompositeAggregate}(x)\rightarrow\exists^{=1}(y)(\mathtt{Part}(x,y)))$\\
 $\forall(x)(\mathtt{CompositeAggregate}(x)\rightarrow\exists^{\leq 1}(y)(\mathtt{Part}(y,x)))$
\end{tabular}

\item relation $(0..*,1)$

\begin{tabular}{>{\footnotesize}p{13cm}}
 $\forall(x,y)((\mathtt{CompositeAggregate}(x)\wedge\mathtt{Whole}(x,y))\rightarrow\mathtt{ObjectType}(y))$\\
 $\forall(x)(\mathtt{CompositeAggregate}(x)\rightarrow\exists^{=1}(y)(\mathtt{Whole}(x,y)))$
\end{tabular}

\item  first textual constraint

\begin{tabular}{>{\footnotesize}p{13cm}}
  $\forall(x,y)(\mathtt{Compatible}(x,y)\rightarrow$\\
\hspace{1.8cm}  $((\mathtt{ValueProperty}(x)\wedge\mathtt{ValueProperty}(y))\vee$\\
\hspace{1.8cm}  $(\mathtt{DataType}(x)\wedge\mathtt{DataType}(y))\vee$\\
\hspace{1.8cm}  $(\mathtt{ObjectType}(x)\wedge\mathtt{ObjectType}(y))\vee$\\
\hspace{1.8cm}  $(\mathtt{Role}(x)\wedge\mathtt{Role}(y))\vee$\\
\hspace{1.8cm}  $(\mathtt{Relationship}(x)\wedge\mathtt{Relationship}(y))))$\\
  $\forall(x,y)((\mathtt{Compatible}(x,y)\wedge\mathtt{Role}(x))\rightarrow\exists(v,w,s,t)(\mathtt{RolePlaying}(x,v,w)\wedge$\\
\hspace{1.8cm}  $\mathtt{RolePlaying}(y,s,t)\wedge\mathtt{Compatible}(w,t)))^*$\\
  $\forall(x,y)((\mathtt{Compatible}(x,y)\wedge\mathtt{Relationship}(x))\rightarrow$\\
\hspace{1.8cm}  $((\exists^{=n}(z)(\mathtt{Contains}(x,z))\leftrightarrow\exists^{=n}(z)(\mathtt{Contains}(y,z)))\wedge$\\
\hspace{1.8cm}  $(\exists(z,v)(\mathtt{Contains}(x,z)\wedge\mathtt{Contains}(y,w)\wedge\mathtt{Compatible}(z,w)))))^*$\\
  $\forall(x,y,z)((\mathtt{Subsumption}(x)\wedge\mathtt{Sub}(x,y)\wedge\mathtt{Super}(x,z))\rightarrow\mathtt{Compatible}(y,z))^*$
  \end{tabular}

\noindent
 Here $n$ is any natural number from 2 up to the maximum arity of a relationship in the model. Therefore the formula is not second order.

\end{itemize}

\subsubsection{Attributes and Value Types}

\begin{figure}[h]
\centering
   \includegraphics[width=1.0\textwidth]{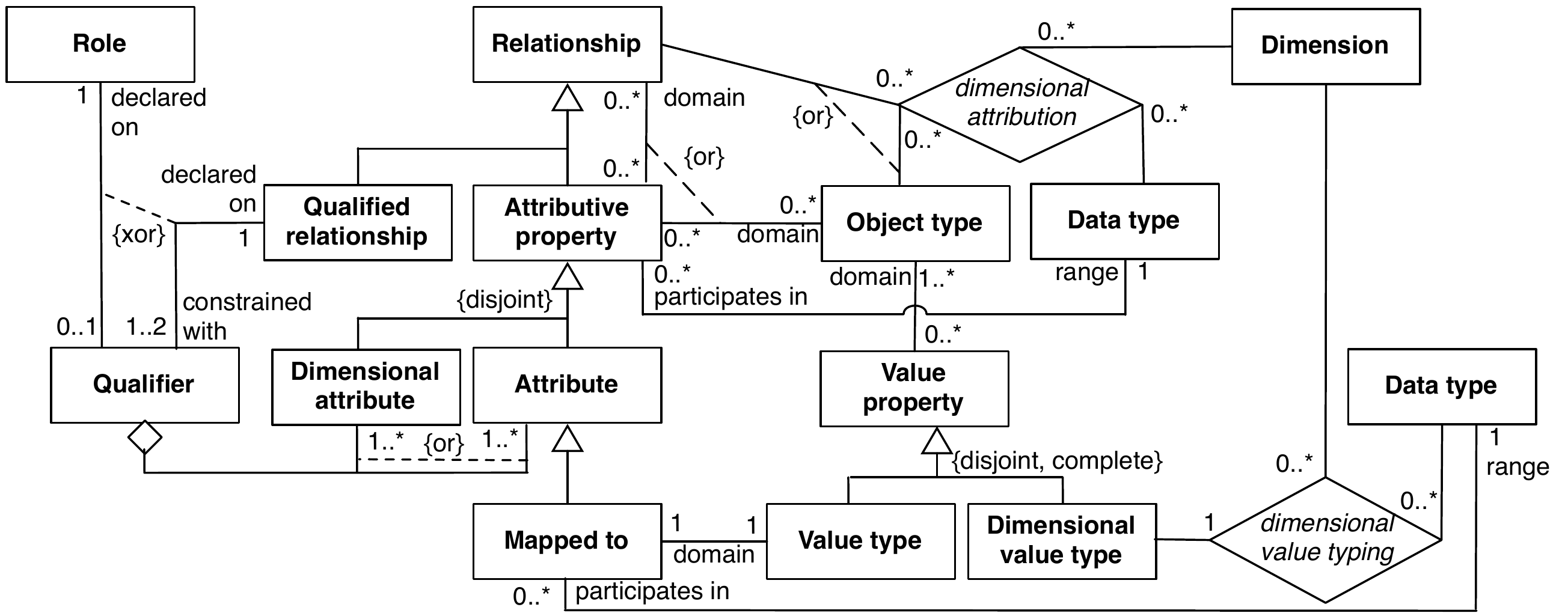}
    \caption{Metamodel fragment for value properties and simple attributes; {\sf Dimensional attribute} is reified version of the ternary relation {\sf dimensional attribution}, and likewise for {\sf Dimensional value type} and {\sf dimensional value typing}.}
    \label{fig:attvt}
\end{figure}

The metamodel fragment concerning attributes and ORM's counterpart, value types, is shown Figure~\ref{fig:attvt} and has the following logic-based reconstruction.

\paragraph{Formalization of Attributes and Value Types}

\begin{itemize}
 \item relation $(0..1,1)$, relation $(1..2,1)$, xor  

\begin{tabular}{>{\footnotesize}p{13cm}}
  $\forall(x,y)(\mathtt{DeclaredOn}(x,y)\rightarrow((\mathtt{Qualifier}(x)\wedge\mathtt{Role}(y))\vee$\\
\hspace{1.8cm}  $(\mathtt{Qualifier}(x)\wedge\mathtt{QualifiedRelationship}(y))\vee$\\
\hspace{1.8cm}  $(\mathtt{MandatoryConstraint}(x)\wedge\mathtt{Role}(y))\vee$\\
\hspace{1.8cm}  $(\mathtt{InternalUniquenessConstraint}(x)\wedge\mathtt{Role}(y))\vee$\\
\hspace{1.8cm}  $(\mathtt{ExternalIdentification}(x)\wedge\mathtt{Relationship}(y))\vee$\\
\hspace{1.8cm}  $(\mathtt{IdentificationConstraint}(x)\wedge\mathtt{ValueProperty}(y))\vee$\\
\hspace{1.8cm}  $(\mathtt{IdentificationConstraint}(x)\wedge\mathtt{AttributiveProperty}(y))\vee$\\       
\hspace{1.8cm}  $(\mathtt{RelationshipEquality}(x)\wedge\mathtt{Relationship}(y))\vee$\\
\hspace{1.8cm}  $(\mathtt{DisjointRelationships}(x)\wedge\mathtt{Relationship}(y))\vee$\\
\hspace{1.8cm}  $(\mathtt{RoleEquality}(x)\wedge\mathtt{Role}(y))\vee$\\
\hspace{1.8cm}  $(\mathtt{DisjointRoles}(x)\wedge\mathtt{Role}(y))\vee$\\
\hspace{1.8cm}  $(\mathtt{DisjointEntities}(x)\wedge\mathtt{Subsumption}(y))\vee$\\
\hspace{1.8cm}  $(\mathtt{ObjectTypeEquality}(x)\wedge\mathtt{ObjectType}(y))\vee$\\
\hspace{1.8cm}  $(\mathtt{CompletenessConstraint}(x)\wedge\mathtt{Subsumption}(y)))   )$\\
  $\forall(x)(\mathtt{Qualifier}(x)\rightarrow\exists^{=1}(y)(\mathtt{DeclaredOn}(x,y)\wedge
(\mathtt{Role}(y)\vee\mathtt{QualifiedRelationship}(y))))$\\
  $\forall(x)(\mathtt{Role}(x)\rightarrow\exists^{\leq 1}(y)(\mathtt{Qualifier}(y)\wedge\mathtt{DeclaredOn}(y,x)))$\\
  $\forall(x)(\mathtt{QualifiedRelationship}(x)\rightarrow(\exists^{\geq 1}(y)(\mathtt{Qualifier}(y)\wedge\mathtt{DeclaredOn}(y,x))\wedge$\\
  \hspace{1.8cm} $\exists^{\leq 2}(y)(\mathtt{Qualifier}(y)\wedge\mathtt{DeclaredOn}(y,x))))$  \\
  $\forall(x,y,z)((\mathtt{DeclaredOn}(x,y)\wedge\mathtt{DeclaredOn}(x,z)\wedge\mathtt{Qualifier}(x)\wedge\mathtt{Role}(y))\rightarrow
\mathtt{Role}(z)))^*$\\
  $\forall(x,y,z)((\mathtt{DeclaredOn}(x,y)\wedge\mathtt{DeclaredOn}(x,z)\wedge\mathtt{Qualifier}(x)\wedge$\\
\hspace{1.8cm}  $\mathtt{QualifiedRelationship}(y))\rightarrow\mathtt{QualifiedRelationship}(z)))^*$
\end{tabular}

\item composition $(0..*, 1..*)$, or 

\begin{tabular}{>{\footnotesize}p{13cm}}
$\forall(x,y)(\mathtt{HasComponent}(x,y)\rightarrow(\mathtt{Qualifier}(x)\wedge
(\mathtt{DimensionalAttribute}(y)\vee\mathtt{Attribute}(y))))$\\
$\forall(x)(\mathtt{Qualifier}(x)\rightarrow\exists^{\geq 1}(y)(\mathtt{HasComponent}(x,y)))$
\end{tabular}

\item relation $(0..*, 0..*)$

\begin{tabular}{>{\footnotesize}p{13cm}}
$\forall(x,y)(\mathtt{Domain}(x,y)\rightarrow((\mathtt{AttributiveProperty}(x)\wedge\mathtt{Relationship}(y))\vee$\\
\hspace{1.8cm}$(\mathtt{AttributiveProperty}(x)\wedge\mathtt{ObjectType}(y))\vee
(\mathtt{MappedTo}(x)\wedge\mathtt{ValueType}(y))\vee$\\
\hspace{1.8cm}$(\mathtt{ValueProperty}(x)\wedge\mathtt{ObjectType}(y))))$
\end{tabular}

\item relation $(0..*, 1)$

\begin{tabular}{>{\footnotesize}p{13cm}}
 $\forall(x,y)(\mathtt{Range}(x,y)\rightarrow((\mathtt{AttributiveProperty}(x)\wedge\mathtt{DataType}(y))\vee$\\
\hspace{1.8cm}$(\mathtt{MappedTo}(x)\wedge\mathtt{DataType}(y))))$\\
 $\forall(x)(\mathtt{AttributiveProperty}(x)\rightarrow\exists^{=1}(y)(\mathtt{DataType}(y)\wedge\mathtt{Range}(x,y)))$
\end{tabular}

\item relation $(0..*, 1..*)$

\begin{tabular}{>{\footnotesize}p{13cm}}
$\forall(x)(\mathtt{ValueProperty}(x)\rightarrow\exists^{\geq 1}(y)(\mathtt{Domain}(x,y)\wedge\mathtt{ObjectType}(y)))$
\end{tabular}

\item relation $(0..*, 1)$

\begin{tabular}{>{\footnotesize}p{13cm}}
 $\forall(x)(\mathtt{MappedTo}(x)\rightarrow\exists^{=1}(y)(\mathtt{DataType}(y)\wedge\mathtt{Range}(x,y)))$
\end{tabular}

\item ternary relation $(0..*, 0..*, 0..*)$, or 

\begin{tabular}{>{\footnotesize}p{13cm}}
$\forall(x,y,z)(\mathtt{DimensionalAttribution}(x,y,z)\rightarrow(\mathtt{Dimension}(x)\wedge\mathtt{DataType}(y)\wedge$\\
\hspace{1.8cm}$(\mathtt{ObjectType}(z)\vee\mathtt{Relationship}(z)))  )$
\end{tabular}

\noindent
Although this formula has three variables, it can be replaced by an equivalent set of several formula in two variables, thus it is not starred.
This translation is similar to the process of reification described in Section~\ref{sec:owl}, and it is applied to all ternary relations.

\item ternary relation $(0..*, 0..*, 1)$

\begin{tabular}{>{\footnotesize}p{13cm}}
$\forall(x,y,z)(\mathtt{DimensionalValueTyping}(x,y,z)\rightarrow(\mathtt{Dimension}(x)\wedge$\\
\hspace{1.8cm}$\mathtt{DataType}(y)\wedge\mathtt{DimensionalValueType}(z)) )$\\
$\forall(x)(\mathtt{DimensionalValueType}(x)\rightarrow\exists^{=1}(y,z)(\mathtt{DimensionalValueTyping}(y,z,x)))$\\
\end{tabular}

\end{itemize}

\subsection{Mandatory Constraints}
\label{sec:mand}

\begin{figure}[h]
\centering
   \includegraphics[width=0.6\textwidth]{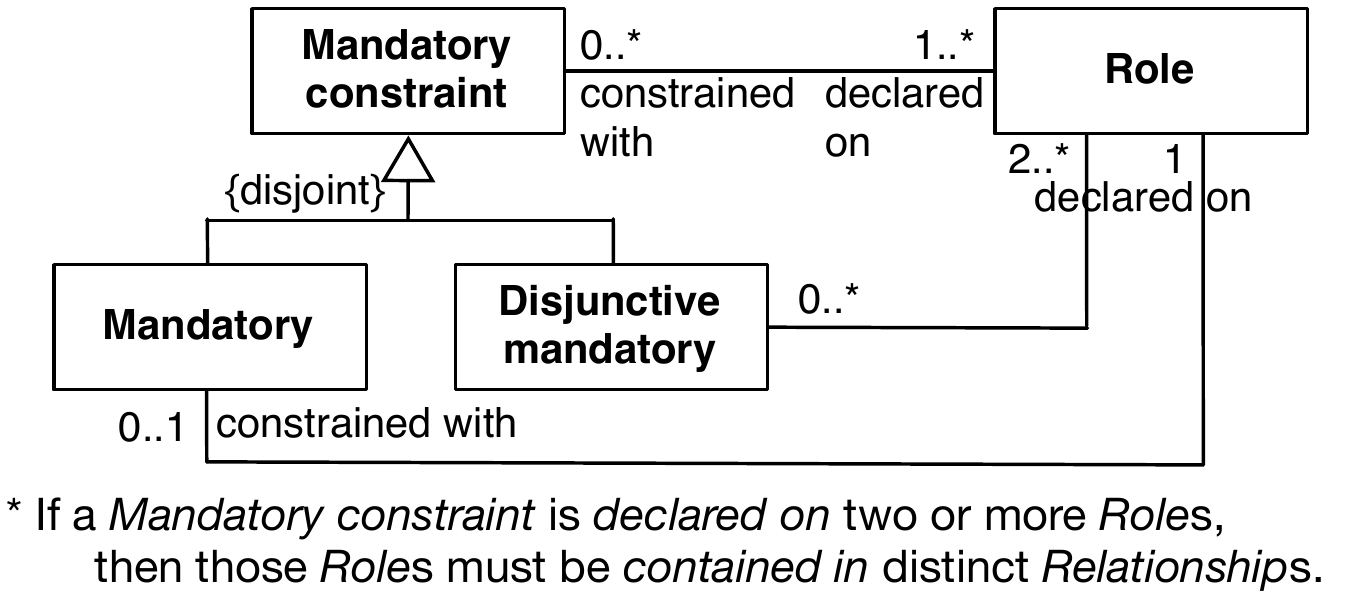}
    \caption{Mandatory constraints.}
    \label{fig:mand}
\end{figure}

Form this section onwards, we turn our focus to constraints that can be represented on the selected conceptual modelling languages. 
The first one is about the different mandatory constraints, as shown in Figure~\ref{fig:mand}.

\paragraph{Formalization of Mandatory Constraints}

\begin{itemize}
\item  relation $(0..*, 1..*)$

\begin{tabular}{>{\footnotesize}p{13cm}}
 $ \forall(x)(\mathtt{MandatoryConstraint}(x)\rightarrow\exists^{\geq 1}(y)(\mathtt{Role}(y)\wedge\mathtt{DeclaredOn}(x,y)))$ 
\end{tabular}

\item relation $(0..*, 2..*)$

\begin{tabular}{>{\footnotesize}p{13cm}}
$\forall(x)(\mathtt{DisjunctiveMandatory}(x)\rightarrow\exists^{\geq 2}(y)(\mathtt{Role}(y)\wedge\mathtt{DeclaredOn}(x,y)))$
\end{tabular}

\item relation $(0..1, 1)$

\begin{tabular}{>{\footnotesize}p{13cm}}
$\forall(x)(\mathtt{Mandatory}(x)\rightarrow\exists^{=1}(y)(\mathtt{Role}(y)\wedge\mathtt{DeclaredOn}(x,y)))$\\
$\forall(x)(\mathtt{Role}(x)\rightarrow\exists^{\leq 1}(y)(\mathtt{DeclaredOn}(y,x)\wedge\mathtt{Mandatory}(y)))$
\end{tabular}

\item  constraint on relationships containing roles with the same mandatory constraint, not in diagram

\begin{tabular}{>{\footnotesize}p{13cm}}
$\forall(x,y,z,v,w)((\mathtt{DeclaredOn}(x,y)\wedge\mathtt{DeclaredOn}(x,z)\wedge\mathtt{MandatoryConstraint}(x)\wedge$\\
\hspace{1.8cm}$\mathtt{Role}(y)\wedge\mathtt{Role}(z)\wedge\mathtt{Contains}(w,y)\wedge\mathtt{Contains}(v,z)\wedge$\\
\hspace{1.8cm}$\mathtt{Relationship}(v)\wedge\mathtt{Relationship}(w))\rightarrow\neg(w=v))^*$
\end{tabular}

\end{itemize}

\subsection{Uniqueness Constraints}

\begin{figure}[h]
\centering
   \includegraphics[width=0.68\textwidth]{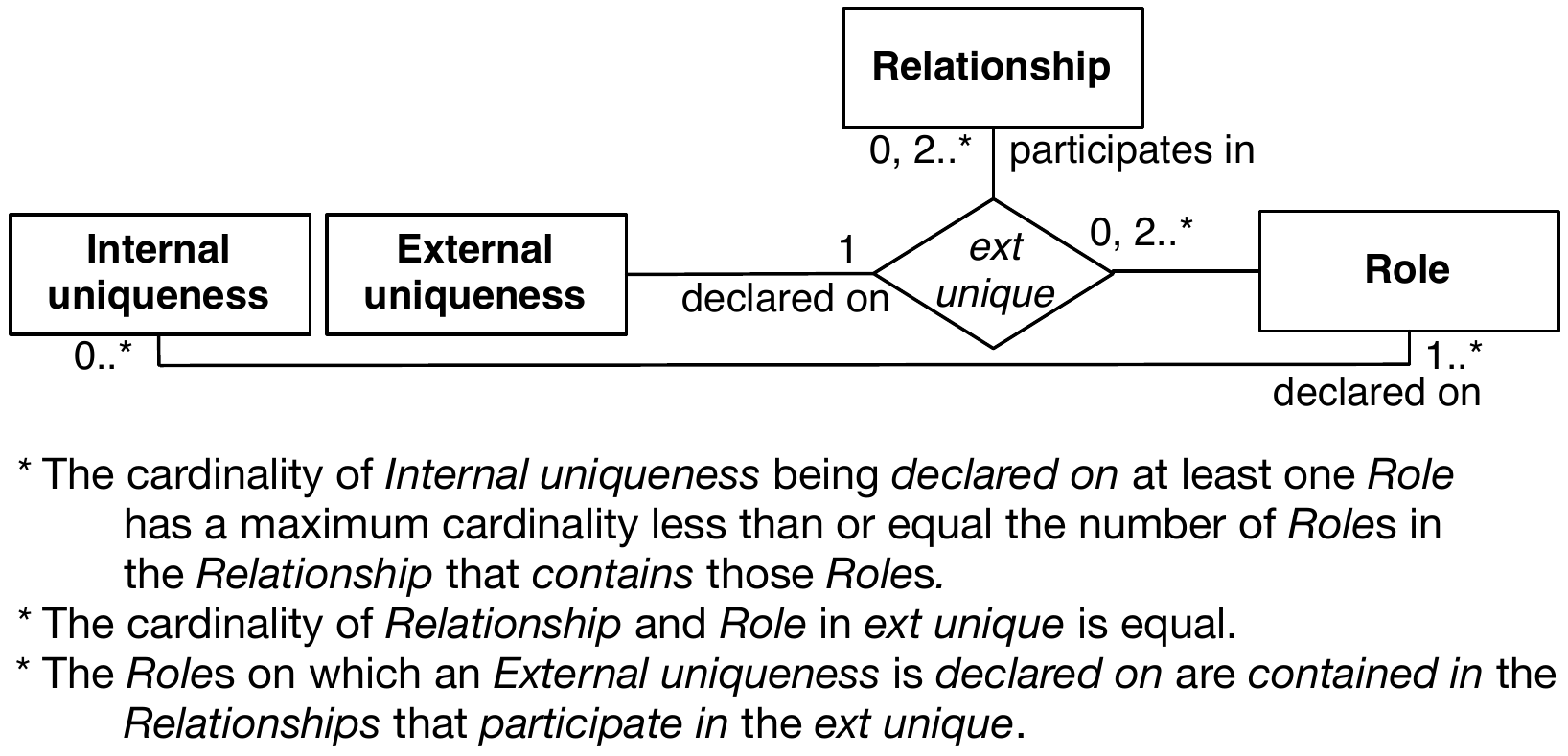}
    \caption{Uniqueness constraints.} 
    \label{fig:unique}
\end{figure}

The second core constraints in the languages is uniqueness, as shown in Figure~\ref{fig:unique}, which is formalised as follows.

\paragraph{Formalization of Uniqueness Constraints}

\begin{itemize}
 \item relation $(0..*,1..*)$

\begin{tabular}{>{\footnotesize}p{13cm}}
 $\forall(x)(\mathtt{InternalUniquenessConstraint}(x)\rightarrow\exists^{\geq 1}(y)(\mathtt{Role}(y)\wedge\mathtt{DeclaredOn}(x,y)))$
\end{tabular}

\item ternary relation $(1, 0;2..*, 0;2..*)$, third textual constraint

\begin{tabular}{>{\footnotesize}p{13cm}}
 $\forall(x,y,z)(\mathtt{ExtUnique}(x,y,z)\rightarrow(\mathtt{ExternalUniquenessConstraint}(x)\wedge$\\
 \hspace{1.8cm}$\mathtt{Role}(y)\wedge\mathtt{Relationship}(z)\wedge\mathtt{Contains}(z,y)))^*$\\
 $\forall(x)(\mathtt{ExternalUniquenessConstraint}(x)\rightarrow\exists^{=1}(y,z)(\mathtt{ExtUnique}(x,y,z)))$\\
 $\forall(x)(\mathtt{Role}(x)\rightarrow(\exists^{=0}(y,z)(\mathtt{ExternalUnique}(y,x,z))\vee$\\
 \hspace{1.8cm}$\exists^{\geq 2}(y,z)(\mathtt{ExternalUnique}(y,x,z))))$\\
 $\forall(x)(\mathtt{Relationship}(x)\rightarrow(\exists^{=0}(y,z)(\mathtt{ExternalUnique}(y,z,x))\vee$\\
 \hspace{1.8cm}$\exists^{\geq 2}(y,z)(\mathtt{ExternalUnique}(y,z,x))))$ 
 \end{tabular}

\item first textual constraint

\begin{tabular}{>{\footnotesize}p{13cm}}
 $\forall(x,y,z)((\mathtt{InternalUniquenessConstraint}(x)\wedge\mathtt{Role}(y)\wedge\mathtt{DeclaredOn}(x,y)\wedge$\\
 \hspace{1.8cm}$\mathtt{Relationship}(z)\wedge\mathtt{Contains}(z,y))\rightarrow $\\
 \hspace{1.8cm}$(\exists^{=n}(w)(\mathtt{Contains}(z,w))\leftrightarrow\exists^{\leq n}(v)(\mathtt{DeclaredOn}(x,v))))^*$
\end{tabular}

\noindent
This is necessary first order since $n$ is any natural number up to the maximum arity of any relationships in the model.

\item second textual constraint

\begin{tabular}{>{\footnotesize}p{13cm}}
 $\forall(x)(\mathtt{ExternalUniquenessConstraint}(x)\rightarrow(\exists^{=n}(y)\exists(z)(\mathtt{ExternalUnique}(x,y,z)$\\
 \hspace{1.8cm}$\leftrightarrow\exists^{=n}(z)\exists(y)\mathtt{ExternalUnique}(x,y,z))))^* $
\end{tabular}

\noindent
This is necessary first order since $n$ is any natural number up to the maximum arity of a relationships in the model.

\end{itemize}

\subsection{Identification Constraints}

\begin{figure}[h]
\centering
   \includegraphics[width=1.0\textwidth]{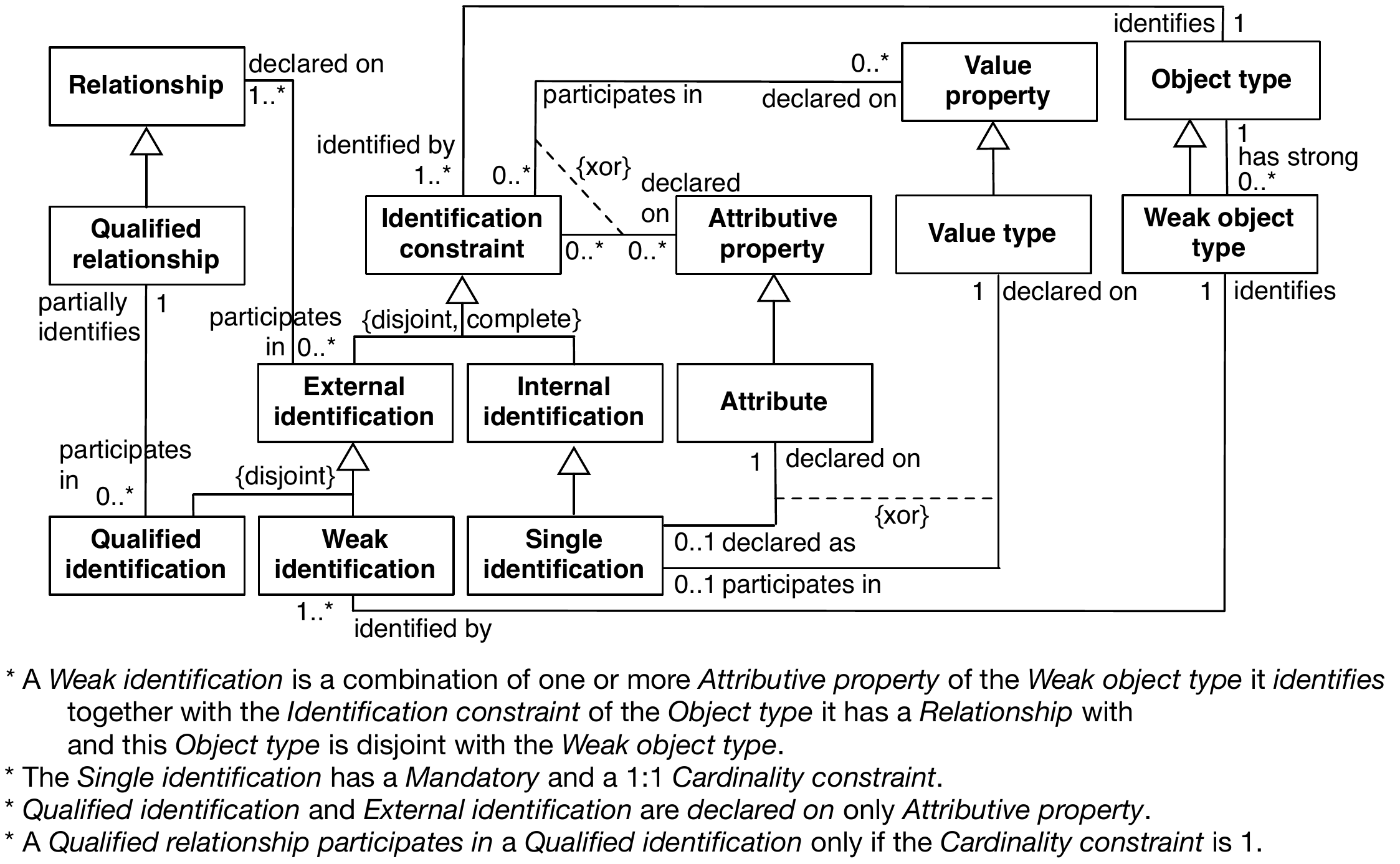}
    \caption{Partial representation of identification constraints.} 
    \label{fig:id}
\end{figure}

The metamodel fragment concerning identifiers (Figure~\ref{fig:id}), while comprehensive in its treatment, does not deal with definitions of identity (see \cite{Keet11keys} for a brief discussion on this topic). How identification is dealt with in the languages is formalised as follows.

\paragraph{Formalization of Identification Constraints}

\begin{itemize}
\item relation $(0..*, 1)$

\begin{tabular}{>{\footnotesize}p{13cm}}
 $\forall(x,y)(\mathtt{PartiallyIdentifies}(x,y)\rightarrow$\\
 \hspace{1.8cm}$(\mathtt{QualifiedIdentification}(x)\wedge\mathtt{QualifiedRelationship}(y))     )$\\
 $\forall(x)(\mathtt{QualifiedIdentification}(x)\rightarrow\exists^{=1}(y)(\mathtt{PartiallyIdentifies}(x,y))   ) $
\end{tabular}

\item relation $(0..*, 1..*)$
 
\begin{tabular}{>{\footnotesize}p{13cm}}
 $\forall(x)(\mathtt{ExternalIdentification}(x)\rightarrow\exists^{\geq 1}(y)(\mathtt{DeclaredOn}(x,y))  $
 \end{tabular}

 \item relation $(1..*, 1)$

\begin{tabular}{>{\footnotesize}p{13cm}}
  $\forall(x,y)(\mathtt{Identifies}(x,y)\rightarrow(\mathtt{IdentificationConstraint}(x)\wedge\mathtt{ObjectType}(y)))$\\
  $\forall(x)(\mathtt{IdentificationConstraint}(x)\rightarrow\exists^{=1}(y)(\mathtt{Identifies}(x,y)))$\\
  $\forall(x)(\mathtt{ObjectType}(x)\rightarrow\exists^{\geq 1}(y)(\mathtt{Identifies}(y,x)))$
 \end{tabular}

 \item relation $(0..*, 0..*)$, xor, mandatory (not in diagram)

\begin{tabular}{>{\footnotesize}p{13cm}}
  $\forall(x,y,z)((\mathtt{DeclaredOn}(x,y)\wedge\mathtt{DeclaredOn}(x,z)\wedge$\\
  \hspace{1.8cm}$\mathtt{IdentificationConstraint}(x)\wedge\mathtt{ValueProperty}(y))\rightarrow$\\
  \hspace{1.8cm}$\mathtt{ValueProperty}(z))^*$\\
  $\forall(x,y,z)((\mathtt{DeclaredOn}(x,y)\wedge\mathtt{DeclaredOn}(x,z)\wedge$\\
  \hspace{1.8cm}$\mathtt{IdentificationConstraint}(x)\wedge\mathtt{AttributiveProperty}(y))\rightarrow$\\
  \hspace{1.8cm}$\mathtt{AttributiveProperty}(z))^*$\\
  $\forall(x)(\mathtt{IdentificationConstraint}(x)\rightarrow\exists(y)(\mathtt{DeclaredOn}(x,y)))$
 \end{tabular}

 \item relation $(0..1, 1)$ 

\begin{tabular}{>{\footnotesize}p{13cm}}
  $\forall(x,y)((\mathtt{DeclaredOn}(x,y)\wedge\mathtt{SingleIdentification}(x))\rightarrow
  (\mathtt{Attribute}(y)\vee\mathtt{ValueType}(y)))$\\
  $\forall(x)(\mathtt{SingleIdentification}(x)\rightarrow\exists^{=1}(y)(\mathtt{DeclaredOn}(x,y))$\\
  $\forall(x)((\mathtt{Attribute}(x)\vee\mathtt{ValueType}(x))\rightarrow\exists^{\leq 1}(y)(\mathtt{DeclaredOn}(y,x)\wedge$\\
  \hspace{1.8cm}$\mathtt{SingleIdentification}(y)))$
\end{tabular}

\noindent
Xor is inherited from previous set of formula.

\item relation $(1..*, 1)$

\begin{tabular}{>{\footnotesize}p{13cm}}
$\forall(x,y)((\mathtt{Identifies}(x,y)\wedge\mathtt{WeakIdentification}(x))\rightarrow\mathtt{WeakObjectType}(y))$\\
$\forall(x)(\mathtt{WeakIdentification}(x)\rightarrow\exists^{= 1}(y)(\mathtt{Identifies}(x,y)))$\\
$\forall(x)(\mathtt{WeakObjectType}(x)\rightarrow\exists^{\geq 1}(y)(\mathtt{WeakIdentification}(y)\wedge$\\
\hspace{1.8cm}$\mathtt{Identifies}(y,x)))$
\end{tabular}

\item relation $(0..*, 1)$

\begin{tabular}{>{\footnotesize}p{13cm}}
$\forall(x,y)(\mathtt{HasStrong}(x,y)\rightarrow(\mathtt{WeakObjectType}(x)\wedge\mathtt{ObjectType}(y))) $\\
$\forall(x)(\mathtt{WeakObjectType}(x)\rightarrow\exists^{= 1}(y)(\mathtt{HasStrong}(x,y)))$
\end{tabular}

\item first textual constraint

\begin{tabular}{>{\footnotesize}p{13cm}}
$\forall(x,y,z)((\mathtt{WeakIdentification}(x)\wedge\mathtt{Identifies}(x,y)\wedge\mathtt{DeclaredOn}(x,z))\rightarrow$\\
\hspace{1.8cm}$((\mathtt{AttributiveProperty}(z)\wedge\mathtt{Domain}(z,y))\vee$\\
\hspace{1.8cm}$\exists^{=1}(v,w,s)(\mathtt{IdentificationConstraint}(v)\wedge\mathtt{Identifies}(v,w)\wedge$\\
\hspace{1.8cm}$\mathtt{Relationship}(s)\wedge\mathtt{DeclaredOn}(v,z)\wedge\mathtt{Participant}(s,w)\wedge
\mathtt{Participant}(s,y))))^*$\\
$\forall(x,y,z,v,w,s,t)((\mathtt{WeakIdentification}(x)\wedge
\mathtt{Identifies}(x,y)\wedge\mathtt{DeclaredOn}(x,s)\wedge$\\
\hspace{1.8cm}$\mathtt{IdentificationConstraint}(v,w)\wedge\mathtt{Identifies}(v,t)\wedge$\\
\hspace{1.8cm}$\mathtt{DeclaredOn}(v,s))\rightarrow\neg(y=t))^*$
\end{tabular}

\item second textual constraint

\begin{tabular}{>{\footnotesize}p{13cm}}
$\forall(x,y,z)((\mathtt{SingleIdentification}(x)\wedge\mathtt{Identifies}(x,y)\wedge\mathtt{DeclaredOn}(x,z))\rightarrow$\\
\hspace{1.8cm}$\exists(v)(\mathtt{CardO}(z,y,v)\wedge\mathtt{MinimumCardinality}(v,1)\wedge
\mathtt{MaximumCardinality}(v,1)))^*$\\
\end{tabular}

\item third textual constraint

\begin{tabular}{>{\footnotesize}p{13cm}}
$\forall(x,y)((\mathtt{QualifiedIdentification}(x)\wedge\mathtt{DeclaredOn}(x,y))\rightarrow
\mathtt{AttributiveProperty}(y))$\\
$\forall(x,y)((\mathtt{ExternalIdentification}(x)\wedge\mathtt{DeclaredOn}(x,y))\rightarrow
\mathtt{AttributiveProperty}(y))$
\end{tabular}

\item fourth textual constraint

\begin{tabular}{>{\footnotesize}p{13cm}}
$\mathtt{\forall(x,y)(\mathtt{PartiallyIdentifies}(x,y)}\rightarrow\exists(z,v,w)(\mathtt{RolePlaying}(z,v,w)\wedge$\\
\hspace{1.8cm}$\mathtt{Contains}(y,z)\wedge\mathtt{MinimumCardinality}(v,1)\wedge
\mathtt{MaximumCardinality}(v,1)))^*$
\end{tabular}
\end{itemize}

\subsection{Cardinality Constraints}

\begin{figure}[ht]
\centering
   \includegraphics[width=0.95\textwidth]{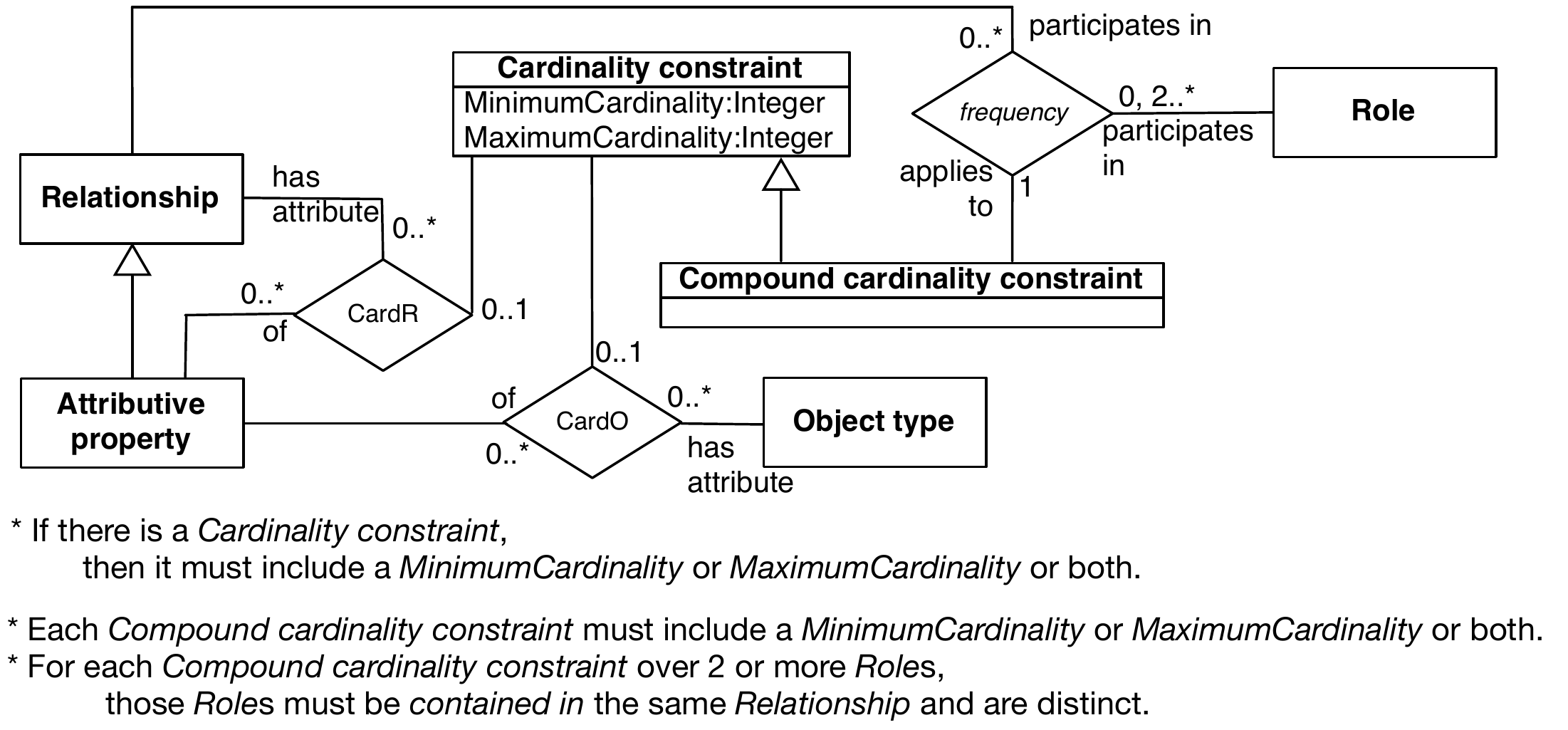}
    \caption{Cardinality constraints on attributes of object types and relationships, and compound cardinality over more than one role in a relationship.} 
    \label{fig:cardAtt}
\end{figure}

The fourth important constraint of the languages are the cardinality constraints; see Figure~\ref{fig:cardAtt}, which is formalised below.

\paragraph{Formalization of Cardinality Constraints}

\begin{itemize}
 \item ternary relationship $(0..*, 0..*, 0..*)$  
 
\begin{tabular}{>{\footnotesize}p{13cm}}
$\forall(x,y,z)(\mathtt{CardR}(x,y,z)\rightarrow(\mathtt{AttributiveProperty}(x)\wedge\mathtt{Relationship}(y)\wedge$\\
\hspace{1.8cm}$\mathtt{CardinalityConstraint}(z)))$
 \end{tabular}

\item ternay relation $(0..*, 0..*, 0..*)$ 

\begin{tabular}{>{\footnotesize}p{13cm}}
$\forall(x,y,z)(\mathtt{CardO}(x,y,z)\rightarrow(\mathtt{AttributiveProperty}(x)\wedge\mathtt{ObjectType}(y)\wedge$\\
\hspace{1.8cm}$\mathtt{CardinalityConstraint}(z)))$
\end{tabular}

\item ternary relation $(0..*, 0;2..*, 1)$

\begin{tabular}{>{\footnotesize}p{13cm}}
$\forall(x,y,z)(\mathtt{Frequency}(x,y,z)\rightarrow(\mathtt{Relationship}(x)\wedge\mathtt{Role}(y)\wedge$\\
\hspace{1.8cm}$\mathtt{CompoundCardinalityConstraint}(z)))$\\
$\forall(x)(\mathtt{CompoundCardinalityConstraint}(x)\rightarrow\exists^{= 1}(y,z)(\mathtt{Frequency}(y,z,x))$\\
$\forall(x)(\mathtt{Role}(x)\rightarrow(\exists^{= 0}(y,z)(\mathtt{Frequency}(y,x,z))\vee$\\
\hspace{1.8cm}$\exists^{\geq 2}(y,z)(\mathtt{Frequency}(y,x,z))))$
\end{tabular}

 \item first and second textual constraint

\begin{tabular}{>{\footnotesize}p{13cm}}
$\forall(x)(\mathtt{CardinalityConstraint}(x)\rightarrow\exists(y)(\mathtt{MaximumCardinality}(x,y)\vee$\\
\hspace{1.8cm}$\mathtt{MinimumCardinality}(x,y)))$
\end{tabular}

 \noindent
 The other part of these constraints are implicit from the formalization of attributes. The second constraint is implicit from the inheritance.

\item isa, third textual constraint 

\begin{tabular}{>{\footnotesize}p{13cm}}
$\forall(x)(\mathtt{CompoundCardinalityConstraint}(x)\rightarrow\mathtt{CardinalityConstraint}(x))$\\
$\forall(x,y,z,v,w)((\mathtt{Frequency}(x,y,z)\wedge\mathtt{Frequency}(v,w,z))\rightarrow$\\
\hspace{1.8cm}$((x=v)\wedge\mathtt{Contains}(x,y)\wedge\mathtt{Contains}(v,w)\wedge\neg(y=w)))^*$
\end{tabular}

\end{itemize}

\subsection{Value type, Role and Attribute Value constraints}

\begin{figure}[ht]
\centering
   \includegraphics[width=0.85\textwidth]{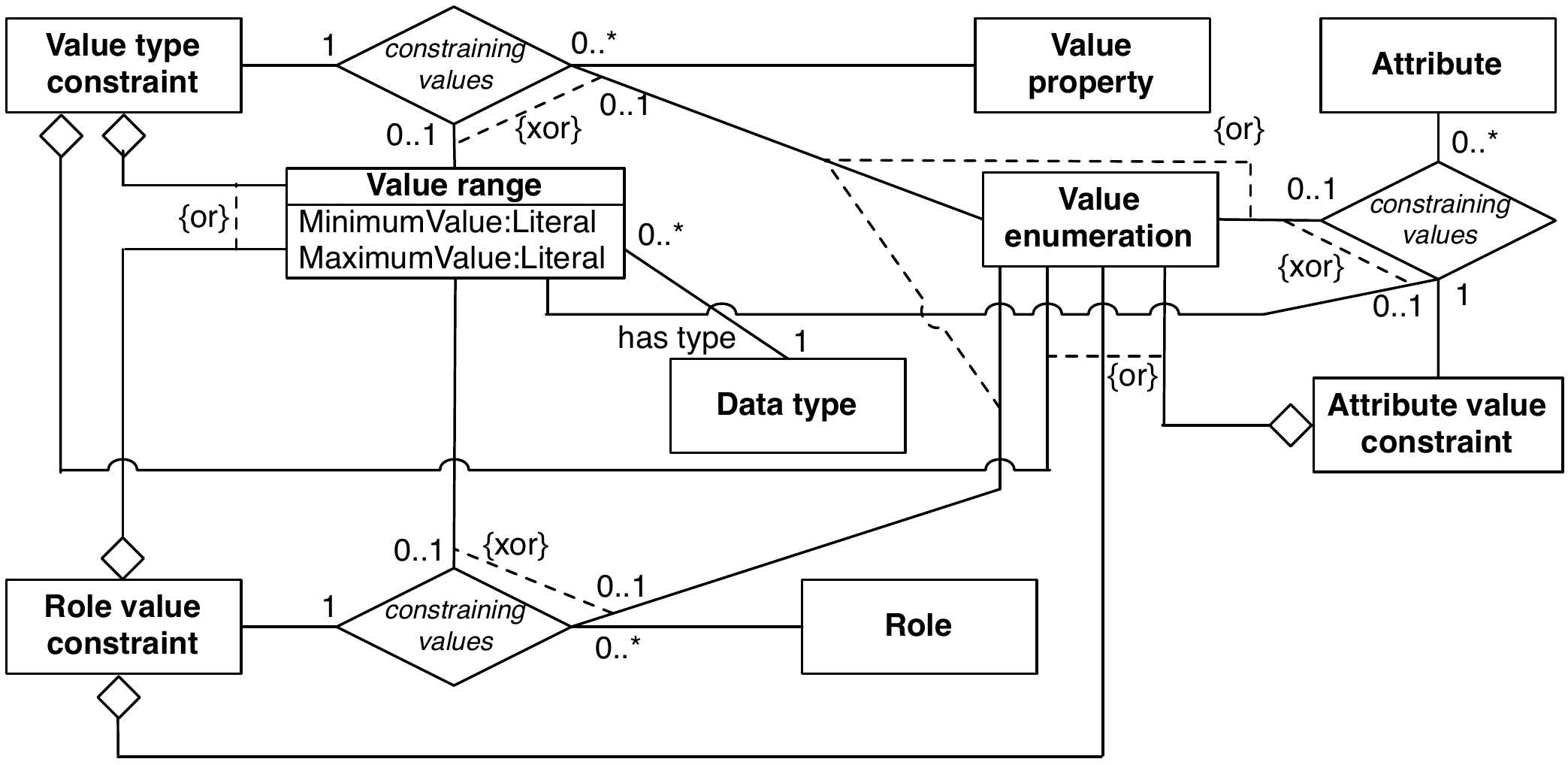}
    \caption{Value type, role, and attribute value constraints.}
    \label{fig:vconT}
\end{figure}

The formalisation of value types and role and attribute constraints is rather elegant compared to the difficulty of drawing that metamodel fragment (Figure~\ref{fig:vconT}), for there are repeating components to it that can be more easily captured in FOL.

\paragraph{Formalization of Value type, Role and Attribute Value Constraints}

\begin{itemize}
 \item ternary relations $(1,0..*,0..*)$, xor

\begin{tabular}{>{\footnotesize}p{13cm}}
$\forall(x,w,z)(\mathtt{ConstrainingValues}(x,y,z)\rightarrow(((\mathtt{ValueTypeConstraint}(x)\wedge$\\
\hspace{1.8cm}$\mathtt{ValueProperty}(y))\vee(\mathtt{RoleValueConstraint}(x)\wedge\mathtt{Role}(y))\vee$\\
\hspace{1.8cm}$(\mathtt{AttributeValueConstraint}(x)\wedge\mathtt{Attribute}(y)))\wedge$\\
\hspace{1.8cm}$(\mathtt{ValueEnumeration}(z)\vee\mathtt{ValueRange}(z))))$\\
$\forall(x)(\mathtt{ValueTypeConstraint}(x)\rightarrow\exists^{= 1}(y,z)(\mathtt{ConstrainingValues}(x,y,z)))$\\
$\forall(x)(\mathtt{RoleValueConstraint}(x)\rightarrow\exists^{= 1}(y,z)(\mathtt{ConstrainingValues}(x,y,z)))$\\
$\forall(x)(\mathtt{AttributeValueConstraint}(x)\rightarrow\exists^{= 1}(y,z)(\mathtt{ConstrainingValues}(x,y,z)))$\\
$\forall(x)(\mathtt{ValueEnumeration}(x)\rightarrow\neg\mathtt{ValueRange}(x))$\\
$\forall(x)(\mathtt{ValueRange}(x)\rightarrow\neg\mathtt{ValueEnumeration}(x))$
\end{tabular}

 \item relation $(0..*,1)$

\begin{tabular}{>{\footnotesize}p{13cm}}
$\forall(x,y)(\mathtt{HasType}(x,y)\rightarrow(\mathtt{ValueRange}(x)\wedge\mathtt{DataType}(y)))$\\
$\forall(x)(\mathtt{ValueRange}(x)\rightarrow\exists^{= 1}(y)(\mathtt{HasType}(x,y)))$
\end{tabular}

\item attributes

\begin{tabular}{>{\footnotesize}p{13cm}}
 $\forall(x,y)((\mathtt{ValueRange}(x)\wedge\mathtt{MinimumValue}(x,y))\rightarrow\mathtt{Literal}(y))$\\
 $\forall(x,y)((\mathtt{ValueRange}(x)\wedge\mathtt{MaximumValue}(x,y))\rightarrow\mathtt{Literal}(y))$\\
 $\forall(x)(\mathtt{ValueRange}(x)\rightarrow\exists^{= 1}(y)(\mathtt{MinimumValue}(x,y)))$\\
 $\forall(x)(\mathtt{ValueRange}(x)\rightarrow\exists^{= 1}(y)(\mathtt{MaximumValue}(x,y)))$ 
\end{tabular}

\noindent
In this figure we assign cardinalities $[1..1]$ to both attributes. The type $\mathtt{Literal}$ is a data type covering all textual
representations of values from simple data types in the model.

\end{itemize}

\subsection{Value comparison constraints}

\begin{figure}[ht]
\centering
   \includegraphics[width=0.8\textwidth]{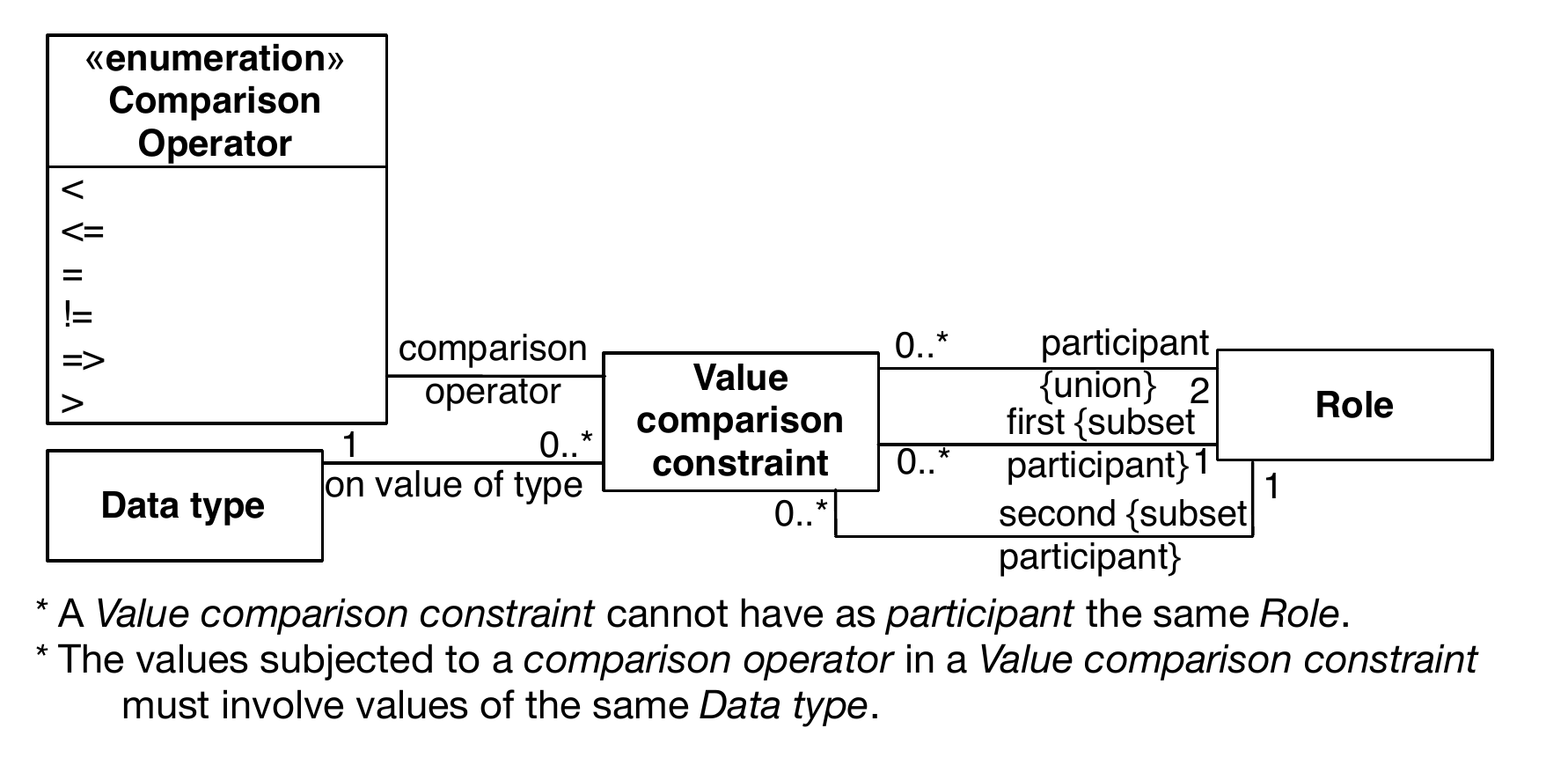}
    \caption{Value comparison constraint; the roles have to be compatible (note: the enumeration already has a {\sf 1..1} constraint).}
    \label{fig:vcomp}
\end{figure}

The metamodel fragment for value comparison constraints is depicted in Figure~\ref{fig:vcomp}. Note that in the formalisation below, a textual representation is chosen for the comparison operators.

\begin{itemize}
 \item relation $(0..*, 2)$  role subset, first textual constraint

\begin{tabular}{>{\footnotesize}p{13cm}}
$\forall(x)(\mathtt{ValueComparisonConstraint}(x)\rightarrow\exists^{=2}(y)(\mathtt{Participant}(x,y)))$\\
$\forall(x,y)(\mathtt{First}(x,y)\rightarrow((\mathtt{ValueComparisonConstraint}(x)\wedge\mathtt{Role}(y))\vee$\\
\hspace{1.8cm}$(\mathtt{RelationshipConstraint}(x)\wedge\mathtt{Role}(y))))$\\
$\forall(x,y)(\mathtt{Second}(x,y)\rightarrow((\mathtt{ValueComparisonConstraint}(x)\wedge\mathtt{Role}(y))\vee$\\
\hspace{1.8cm}$(\mathtt{RelationshipConstraint}(x)\wedge\mathtt{Role}(y))))$\\
$\forall(x,y)(\mathtt{First}(x,y)\rightarrow\mathtt{Participant}(x,y))$\\
$\forall(x,y)(\mathtt{Second}(x,y)\rightarrow\mathtt{Participant}(x,y))$\\
$\forall(x,y)((\mathtt{Participant}(x,y)\wedge\mathtt{ValueComparisonConstraint}(x))\rightarrow$\\
\hspace{1.8cm}$(\mathtt{First}(x,y)\vee\mathtt{Second}(x,y)))$\\
$\forall(x,y)(\neg(\mathtt{First}(x,y)\wedge\mathtt{Second}(x,y)\wedge\mathtt{ValueComparisonConstraint}(x)))$\\
$\forall(x)(\mathtt{ValueComparisonConstraint}(x)\rightarrow\exists^{= 1}(y)(\mathtt{First}(x,y)))$\\
$\forall(x)(\mathtt{ValueComparisonConstraint}(x)\rightarrow\exists^{= 1}(y)(\mathtt{Second}(x,y)))$
\end{tabular}

\item nominals

\begin{tabular}{>{\footnotesize}p{13cm}}
$\forall(x,y)(\mathtt{ComparisonOperatorOf}(x,y)\rightarrow
(\mathtt{ValueComparisonConstraint}(x)\wedge\mathtt{Operator}(y)))$\\
$\forall(x)(\mathtt{ValueComparisonConstraint}(x)\rightarrow\exists^{=1}(y)(\mathtt{ComparisonOperatorOf}(x,y)))$\\
$\forall(x)(\mathtt{Operator}(x)\rightarrow((x=\mathtt{Less})\vee(x=\mathtt{Leq})\vee(x=\mathtt{Eq})\vee$\\
\hspace{1.8cm}$(x=\mathtt{Neq})\vee(x=\mathtt{Geq})\vee(x=\mathtt{Greater})))$\\
$\forall(x)((x=\mathtt{Less})\rightarrow(\mathtt{Operator}(x)\wedge\neg(x=\mathtt{Leq})\wedge\neg(x=\mathtt{Eq})\wedge$\\
\hspace{1.8cm}$\neg(x=\mathtt{Neq})\wedge\neg(x=\mathtt{Geq})\wedge\neg(x=\mathtt{Greater})))$\\
$\forall(x)((x=\mathtt{Leq})\rightarrow(\mathtt{Operator}(x)\wedge\neg(x=\mathtt{Less})\wedge\neg(x=\mathtt{Eq})\wedge$\\
\hspace{1.8cm}$\neg(x=\mathtt{Neq})\wedge\neg(x=\mathtt{Geq})\wedge\neg(x=\mathtt{Greater})))$\\
$\forall(x)((x=\mathtt{Eq})\rightarrow(\mathtt{Operator}(x)\wedge\neg(x=\mathtt{Less})\wedge\neg(x=\mathtt{Leq})\wedge$\\
\hspace{1.8cm}$\neg(x=\mathtt{Neq})\wedge\neg(x=\mathtt{Geq})\wedge\neg(x=\mathtt{Greater})))$\\
$\forall(x)((x=\mathtt{Neq})\rightarrow(\mathtt{Operator}(x)\wedge\neg(x=\mathtt{Less})\wedge\neg(x=\mathtt{Leq})\wedge$\\
\hspace{1.8cm}$\neg(x=\mathtt{Eq})\wedge\neg(x=\mathtt{Geq})\wedge\neg(x=\mathtt{Greater})))$\\
$\forall(x)((x=\mathtt{Geq})\rightarrow(\mathtt{Operator}(x)\wedge\neg(x=\mathtt{Less})\wedge\neg(x=\mathtt{Leq})\wedge$\\
\hspace{1.8cm}$\neg(x=\mathtt{Eq})\wedge\neg(x=\mathtt{Neq})\wedge\neg(x=\mathtt{Greater})))$\\
$\forall(x)((x=\mathtt{Greater})\rightarrow(\mathtt{Operator}(x)\wedge\neg(x=\mathtt{Less})\wedge\neg(x=\mathtt{Leq})\wedge$\\
\hspace{1.8cm}$\neg(x=\mathtt{Eq})\wedge\neg(x=\mathtt{Neq})\wedge\neg(x=\mathtt{Geq})))$
\end{tabular}

\noindent
$\mathtt{Operator}$ is the name assigned to the concept associated with the nominals in Figure~\ref{fig:vcomp}. These nominals are the only operators allowed,
and they are all different.

\item relation $(1, 0..*)$

\begin{tabular}{>{\footnotesize}p{13cm}}
$\forall(x,y)(\mathtt{OnValueOfType}(x,y)\rightarrow(\mathtt{ValueComparisonConstraint}(x)\wedge\mathtt{DataType}(y)))$\\
$\forall(x)(\mathtt{ValueComparisonConstraint}(x)\rightarrow\exists^{= 1}(y)(\mathtt{OnValueOfType}(x,y)))$
\end{tabular}

 \item second textual constraint

\begin{tabular}{>{\footnotesize}p{13cm}}
$\forall(x,y,z)((\mathtt{ValueComparisonConstraint}(x)\wedge\mathtt{First}(x,y)\wedge\mathtt{Second}(x,z))\rightarrow$\\
\hspace{1.8cm}$\exists(u,v,w)(\mathtt{RolePlaying}(y,u,v)\wedge\mathtt{RolePlaying}(z,w,v)
\wedge\mathtt{DataType}(v)))^*$
\end{tabular}
\end{itemize}

\subsection{Role and Relationship constraints}

\begin{figure}[t]
\centering
   \includegraphics[width=0.8\textwidth]{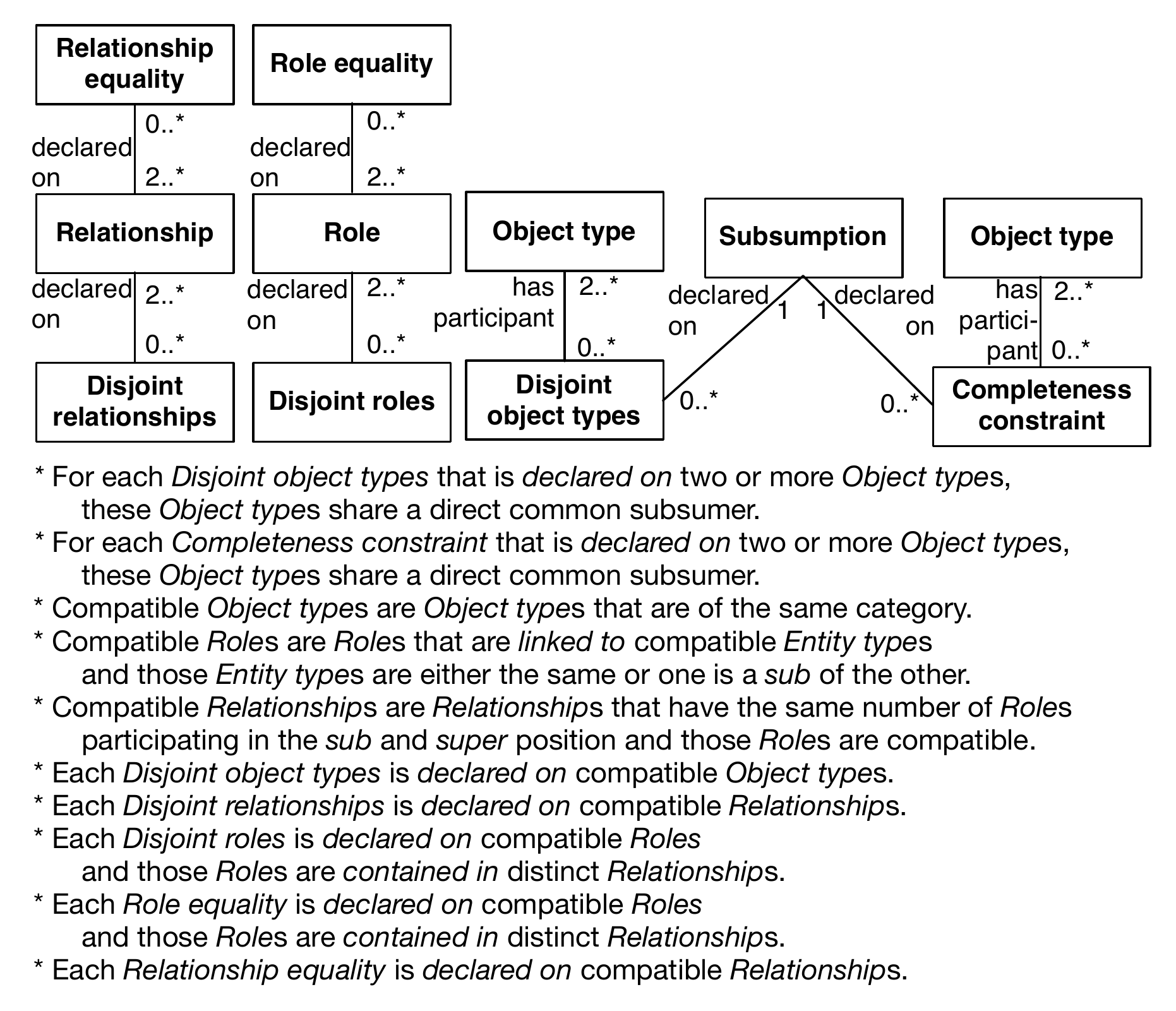}
    \caption{Disjointness and completeness constraints, and role and relationship equality.} 
    \label{fig:disj}
\end{figure}

\begin{figure}[h]
\centering
   \includegraphics[width=0.55\textwidth]{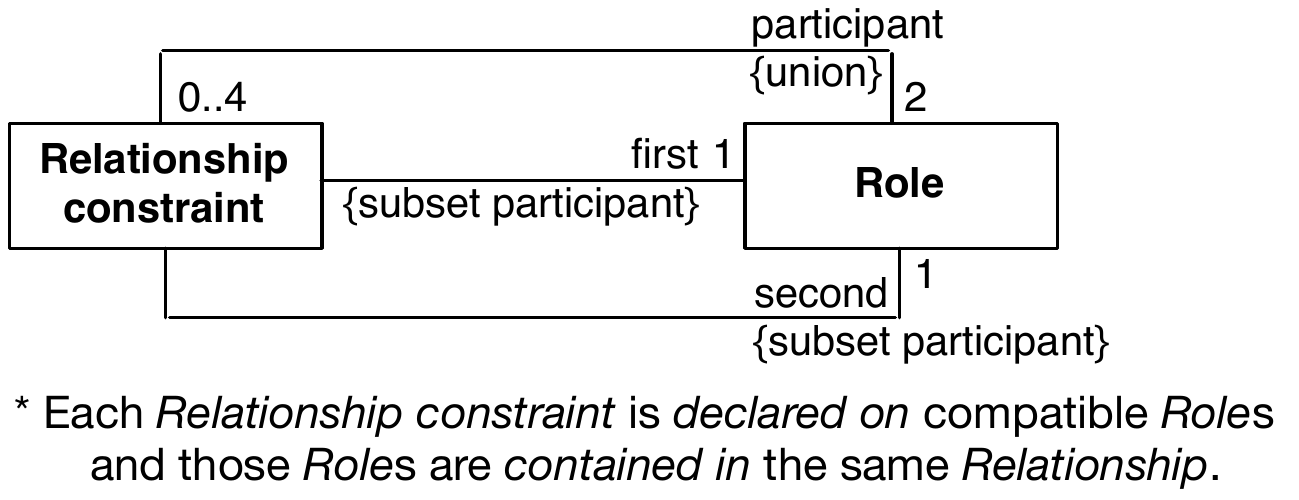}
    \caption{Relationship constraints (see Figure~\ref{fig:constraints} for the hierarchy of relational properties).}
    \label{fig:relcon}
\end{figure}

\begin{figure}[t]
\centering
   \includegraphics[width=0.99\textwidth]{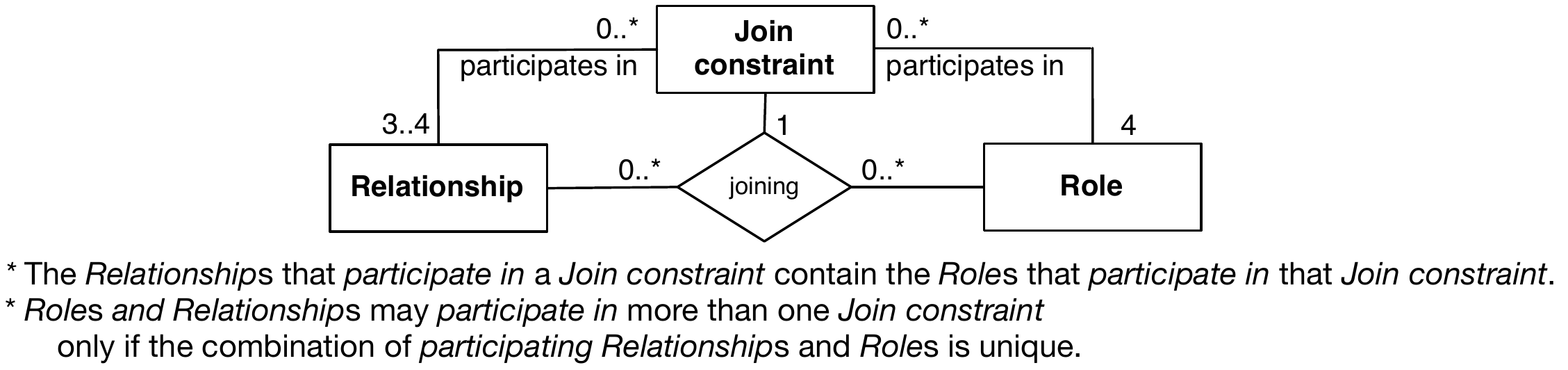}
    \caption{Join constraints.}
    \label{fig:joins}
\end{figure}

Constraints among roles and relationship are shown in Figure~\ref{fig:disj}, and it, with its repetitive list of similar textual constraints are formalised afterward in this section. The Relationship constraints (reflexivity etc.) are presented afterward in Figure~\ref{fig:relcon} and its formalisation. This section closes with the metamodel fragment for ORM's join constraints, depicted in Figure~\ref{fig:joins} and formalised at the end of this section, therewith completing the metamodel and its FOL formalisation.

\paragraph{Formalization of Equality and Disjointness Constraints}

\begin{itemize}
 \item four relations $\mathtt{DeclaredOn}$ $(0..*,  2..*)$

\begin{tabular}{>{\footnotesize}p{13cm}}
  $\forall(x)(\mathtt{RelationshipEquality}(x)\rightarrow\exists^{\geq 2}(y)(\mathtt{DeclaredOn}(x,y)))$\\
  $\forall(x)(\mathtt{DisjointRelationships}(x)\rightarrow\exists^{\geq 2}(y)(\mathtt{DeclaredOn}(x,y)))$\\
  $\forall(x)(\mathtt{RoleEquality}(x)\rightarrow\exists^{\geq 2}(y)(\mathtt{DeclaredOn}(x,y)))$\\
  $\forall(x)(\mathtt{DisjointRoles}(x)\rightarrow\exists^{\geq 2}(y)(\mathtt{DeclaredOn}(x,y)))$
 \end{tabular}

 \item two relations $\mathtt{DeclaredOn}$ $(0..*, 1)$

\begin{tabular}{>{\footnotesize}p{13cm}}
  $\forall(x)(\mathtt{CompletenessConstraint}(x)\rightarrow\exists^{= 1}(y)(\mathtt{DeclaredOn}(x,y)))$\\
  $\forall(x)(\mathtt{DisjointEntities}(x)\rightarrow\exists^{= 1}(y)(\mathtt{DeclaredOn}(x,y)))$\\
\end{tabular}

 \item two relations $\mathtt{hasParticipant}$ $(0..*, 2..*)$

\begin{tabular}{>{\footnotesize}p{13cm}}
  $\forall(x,y)(\mathtt{HasParticipant}(x,y)\rightarrow
  ((\mathtt{CompletenessConstraint}(x)\vee\mathtt{DisjointEntities}(x))\wedge$\\
  \hspace{1.8cm}$\mathtt{Entity}(y)\wedge \neg\mathtt{QualifiedRelationship}(y)\wedge\neg\mathtt{Subsumption}(y)\wedge$\\
  \hspace{1.8cm}$\neg\mathtt{AttributiveProperty}(y)\wedge\neg\mathtt{Qualifier}(y)\wedge\neg\mathtt{Constraint}(y) ))$\\
  $\forall(x)(\mathtt{CompletenessConstraint}(x)\rightarrow\exists^{\geq 2}(y)(\mathtt{HasParticipant}(x,y)))$\\
  $\forall(x)(\mathtt{DisjointEntities}(x)\rightarrow\exists^{\geq 2}(y)(\mathtt{HasParticipant}(x,y)))$  
 \end{tabular}

 \item textual constraints 

 \begin{tabular}{>{\footnotesize}p{13cm}}
  $\forall(x,y)((\mathtt{DisjointEntities}(x)\wedge\mathtt{HasParticipant}(x,y))\rightarrow$\\
  \hspace{1.8cm}$\exists(z)(\mathtt{DeclaredOn}(x,z)\wedge\mathtt{Subsumption}(z)\wedge\mathtt{Sub}(z,y)))^*$\\
  $\forall(x,y,z)((\mathtt{DisjointEntities}(x)\wedge\mathtt{DeclaredOn}(x,y)\wedge\mathtt{Sub}(y,z))\rightarrow
\mathtt{HasParticipant}(x,z))^*$\\
  $\forall(x,y)((\mathtt{CompletenessConstraint}(x)\wedge\mathtt{HasParticipant}(x,y))\rightarrow$\\
  \hspace{1.8cm}$\exists(z)(\mathtt{DeclaredOn}(x,z)\wedge\mathtt{Subsumption}(z)\wedge\mathtt{Sub}(z,y)))^*$\\
  $\forall(x,y,z)((\mathtt{CompletenessConstraint}(x)\wedge\mathtt{DeclaredOn}(x,y)\wedge\mathtt{Sub}(y,z))\rightarrow$\\
  \hspace{1.8cm}$\mathtt{HasParticipant}(x,z))^*$\\
  $\forall(x,y)(\mathtt{Compatible}(x,y)\rightarrow$\\
  \hspace{1.8cm}$((\mathtt{ValueProperty}(x)\wedge\mathtt{ValueProperty}(y))\vee$\\
  \hspace{1.8cm}$(\mathtt{DataType}(x)\wedge\mathtt{DataType}(y))\vee$\\
  \hspace{1.8cm}$(\mathtt{ObjectType}(x)\wedge\mathtt{ObjectType}(y))\vee$\\
  \hspace{1.8cm}$(\mathtt{Role}(x)\wedge\mathtt{Role}(y))\vee$\\
  \hspace{1.8cm}$(\mathtt{Relationship}(x)\wedge\mathtt{Relationship}(y))))$\\
  $\forall(x,y)((\mathtt{Compatible}(x,y)\wedge\mathtt{Role}(x))\rightarrow$\\
  \hspace{1.8cm}$\exists(v,w,s,t)(\mathtt{RolePlaying}(x,v,w)\wedge\mathtt{RolePlaying}(y,s,t)\wedge
\mathtt{Compatible}(w,t)))^*$\\
  $\forall(x,y)((\mathtt{Compatible}(x,y)\wedge\mathtt{Relationship}(x))\rightarrow$\\
  \hspace{1.8cm}$((\exists^{=n}(z)(\mathtt{Contains}(x,z))\leftrightarrow\exists^{=n}(z)(\mathtt{Contains}(y,z)))\wedge$\\
  \hspace{1.8cm}$(\exists(z,v)(\mathtt{Contains}(x,z)\wedge\mathtt{Contains}(y,w)\wedge\mathtt{Compatible}(z,w)))))^*$\\
    $\forall(x,y,z)((\mathtt{DisjointRelationships}(x)\wedge\mathtt{DeclaredOn}(x,y)\wedge\mathtt{DeclaredOn}(x,z))\rightarrow$\\
    \hspace{1.8cm}$\mathtt{Compatible}(y,z))^*$\\
    $\forall(x,y,z)((\mathtt{DisjointRoles}(x)\wedge\mathtt{DeclaredOn}(x,y)\wedge\mathtt{DeclaredOn}(x,z))\rightarrow
    (\mathtt{Compatible}(y,z)\wedge$\\
    \hspace{1.8cm}$\forall(v,w)((\mathtt{Contains}(v,y)\wedge\mathtt{Contains}(w,z))\rightarrow\neg(y=z))))^*$\\
     $\forall(x,y,z)((\mathtt{RoleEquality}(x)\wedge\mathtt{DeclaredOn}(x,y)\wedge\mathtt{DeclaredOn}(x,z))\rightarrow
    (\mathtt{Compatible}(y,z)\wedge$\\
    \hspace{1.8cm}$\forall(v,w)((\mathtt{Contains}(v,y)\wedge\mathtt{Contains}(w,z))\rightarrow\neg(y=z))))^*$\\
    $\forall(x,y,z)((\mathtt{RelationshipEquality}(x)\wedge\mathtt{DeclaredOn}(x,y)\wedge\mathtt{DeclaredOn}(x,z))\rightarrow$\\
    \hspace{1.8cm}$\mathtt{Compatible}(y,z))^*$
 \end{tabular}

\noindent
This is necessary first order since $n$ is any natural number up to the maximum arity of a relationships in the model.
\end{itemize}

\paragraph{Formalization of Relationship Constraints}

\begin{itemize}
 \item relation $(0..4, 2)$ 
 
\begin{tabular}{>{\footnotesize}p{13cm}}
$\forall(x)(\mathtt{RelationshipConstraint}(x)\rightarrow\exists^{=2}(y)(\mathtt{Participant}(x,y)))$\\
$\forall(x)(\mathtt{Role}(x)\rightarrow\exists^{\leq 4}(y)(\mathtt{Participant}(y,x)))$
\end{tabular}

\item two relations $(0..*, 1)$, subset, union

\begin{tabular}{>{\footnotesize}p{13cm}}
$\forall(x,y)((\mathtt{Participant}(x,y)\wedge\mathtt{RelationshipConstraint}(x))\rightarrow
 (\mathtt{First}(x,y)\vee\mathtt{Second}(x,y)))$\\
$\forall(x,y)(\neg(\mathtt{First}(x,y)\wedge\mathtt{Second}(x,y)\wedge\mathtt{RelationshipConstraint}(x)))$\\
$\forall(x)(\mathtt{RelationshipConstraint}(x)\rightarrow\exists^{= 1}(y)(\mathtt{First}(x,y)))$\\
$\forall(x)(\mathtt{RelationshipConstraint}(x)\rightarrow\exists^{= 1}(y)(\mathtt{Second}(x,y)))$
\end{tabular}

\item textual constraint

\begin{tabular}{>{\footnotesize}p{13cm}}
$\forall(x,y,z)((\mathtt{RelationshipConstraint}(x,y)\wedge\mathtt{RelationshipConstraint}(x,z)\wedge$\\
\hspace{1.8cm}$\mathtt{First}(x,y)\wedge\mathtt{Second}(x,z))\rightarrow(\mathtt{Compatible}(y,z)\wedge$\\
\hspace{1.8cm}$\exists(w)(\mathtt{Contains}(w,y)\wedge\mathtt{Contains}(w,z))))^*$
\end{tabular}
\end{itemize}

\paragraph{Formalization of Join Constraints}

\begin{itemize}
 \item ternary relation $(0..*, 1, 0..*)$

\begin{tabular}{>{\footnotesize}p{13cm}}
$\forall(x,y,z)(\mathtt{Joining}(x,y,z)\rightarrow(\mathtt{Relationship}(x)\wedge\mathtt{Role}(z)\wedge
\mathtt{JoinConstraint}(y)))$\\
$\forall(x)(\mathtt{JoinConstraint}(x)\rightarrow\exists^{= 1}(y,z)(\mathtt{Joining}(y,x,z)))$
 \end{tabular}

\item relation $(3..4, 0..*)$ and relation $(4, 0..*)$

\begin{tabular}{>{\footnotesize}p{13cm}}
$\forall(x,y)(\mathtt{ParticipatesIn}(x,y)\rightarrow(\mathtt{JoinConstraint}(y)\wedge
(\mathtt{Relationship}(x)\vee\mathtt{Role}(x))))$\\
$\forall(x)(\mathtt{JoinConstraint}(x)\rightarrow(\exists^{=3}(y)(\mathtt{Relationship}(y)\wedge\mathtt{ParticipatesIn}(y,x))\vee$\\
\hspace{1.8cm}$\exists^{=4}(y)(\mathtt{Relationship}(y)\wedge\mathtt{ParticipatesIn}(x,y))))$\\
$\forall(x)(\mathtt{JoinConstraint}(x)\rightarrow\exists^{=4}(y)(\mathtt{Role}(y)\wedge\mathtt{ParticipatesIn}(y,x)))$
\end{tabular}

\item first textual constraint

\begin{tabular}{>{\footnotesize}p{13cm}}
$\forall(x,y)((\mathtt{JoinConstraint}(x)\wedge\mathtt{Role}(y)\wedge\mathtt{ParticipatesIn}(y,x))\rightarrow$\\
\hspace{1.8cm}$\exists(z)(\mathtt{Relationship}(z)\wedge\mathtt{Contains}(z,y)\wedge\mathtt{ParticipatesIn}(z,x)))^*$
\end{tabular}
\end{itemize}

\subsection{On the complexity of the FOL formalization}
\label{sec:properties}

The complete FOL formalization of the metamodel is probably undecidable. We have complex formula in five variables, counting quantifiers, and ternary predicates.
Even if we reify ternary relations, the complexity of deciding if a given formula follows from some instantiated metamodel is similar to that of the general
satisfiability problem. Problematic formula belong though to textual constraints of the figures; they are not originated from the graphical language of UML class
diagrams. So if we remove from the metamodel all formula labeled with a ``$^*$'', then we obtain a lighter formalization described in $C^2$, the fragment of function-free FOL with only two variables and counting
quantifiers. This fragment is not only decidable, but also in NEXPTIME, as shown in \cite{pratt2005complexity}. Most description logics are also subsets of 
$C^2$.  

This lighter version of the metamodel is the
basis for the OWL 2 version described in next section.

\section{OWL 2 version of the metamodel}
\label{sec:owl}

Considering that the complete formalisation of the metamodel requires full FOL, hence, using an undecidable language, we also look at representing 
a subset in a decidable language to facilitate the metamodel's use in various applications. A relevant and popular logic language in this context is OWL,
so that, thanks to the tools that can process it, we also can have some  form of verification that our model is consistent (at least for what is
representable in OWL 2 DL). This OWL version of the metamodel (v1.2) is available at \url{http://www.meteck.org/files/ontologies/metamodelCDML.owl}.

The OWL version is larger than the number of entities in the diagrams of the preceding section, and has  99 classes, 
61 object properties, 2 data properties, and overall 657 axioms, and the language used is $\mathcal{SHIQ}(D)$ (data from 
Prot\'eg\'e v4.3's ``ontology metrics'' page), i.e., using features of OWL 2 DL \cite{OWL2rec} and in particular the qualified cardinality. Because the 
diagrams were made in Omnigraffle\footnote{\url{https://www.omnigroup.com/omnigraffle}; last accessed 5-12-2014.}, no UML-in-Omnigraffle to OWL 
tool exists, there are textual constraints in our metamodel that can be captured formally, and other UML-to-OWL tools are partial, we formalised 
the metamodel manually. Further, due to the limitations of OWL on the one hand, and the greater expressiveness regarding other features, there is 
no exact match between the metamodel and the OWL file. We describe the main changes and design decisions in the remainder of this section. 

\paragraph{Missing information.}

The first thing to note is that the ternary and quaternary associations of the metamodel could not be represented in OWL, and thus had to be approximated.
We used the following procedure for this, 
which is similar to procedures mentioned elsewhere (e.g., the ontology design pattern in \cite{Presutti08}):
\begin{compactenum}
\item Add new class {\sf ReifiedNAry}, 
\item If the n-ary was not reified already then add it as subclass of {\sf ReifiedNAry} and append it with an {\sf R}, if present already then add a subsumption to {\sf ReifiedNAry}. 
\item Add n binaries between said class and the others, where the binaries have a naming scheme of the name of the n-ary appended with a number ({\sf 1}, {\sf 2} etc.) to distinguish them from real object properties, and add domain and range classes. 
\item These numbered object properties are not to be used as real object properties elsewhere in the OWL file. 
\item Add an exactly 1 (or other cardinality constraint, depending on the constraint in the model) from the {\sf ReifiedNAryR} to the participating class.
\end{compactenum}
Note that several ternary relations in the metamodel are already reified, since one of their participant has cardinality constraint $1..1$. For these relations,
the above procedure can be simplified.

Other information missing from the OWL file are the textual constraints whose formula are marked with ``$^*$'' in the FOL formalisations, because these constraints
cannot be expressed in OWL. Also, a few other complex constraints, such as the ``The Single identification has a Mandatory constraint on the participating Role and the Relationship that Role is contained in has a 1:1 Cardinality constraint declared on it.'' (see second textual constraint in Figure~\ref{fig:id}), are not included due to OWL's expressiveness limitations.
%

\paragraph{Other modelling artefacts.} Some other representation choices had to be made. We note them here for information and documentation; their rationale is outside the scope of this technical report. 

There are several entities that were not subsumed by anything in particular, such as {\sf MaximumCardinality}, {\sf MinimumCardinality}, {\sf Dimension}, and {\sf ValueRange}, which have now been made an OWL subclass of {\sf Abstract}. This also meant that the four attributes in the metamodel have been `flattened' on the OWL file, i.e., instead of one binary from class to data type with OWL's data property, it uses a separate part relation and then a {\sf hasCValue} and {\sf hasValue}, respectively, to the values. 

The participation of {\sf IdentificationConstraint} is twice {\sf 0..*} in the figure, but that is translated into OWL as ``some (ValueProperty or AttributiveProperty)'', because the constraint has to be declared on something.  

The comparison operators of Figure~\ref{fig:vcomp} are written out in text and represented as subclasses of {\sf Comparison} and is made the disjoint union of its subclasses, and the {\sf ValueComparisonConstraint} is given exactly 1 {\sf Comparison}.  

It has been shown that using OWL 2's new feature of inverse object properties results in better reasoner performance \cite{Keet14ore}. Therefore, we chose not to use object properties `in both directions', but use that instead. For instance {\sf constrainedWith} has not been used in the OWL file, but instead, in Prot\'eg\'e notation, {\sf inverse(declaredOn)}.

We have added {\sf partOf}, {\sf properPartOf}, and {\sf hasPart}. 


\paragraph{Satisfiability and consistency checking.}
We checked the OWL-ized metamodel with the automated reasoner HerMiT v1.3.8 that is shipped with Prot\'eg\'e v4.3. 
The sole useful inference was the detection of an inconsistency of {\sf Asymmetry} in the hierarchy of constraints 
in an earlier version of the metamodel: the $\{${\sf disjoint}, {\sf complete}$\}$ on the object subtypes of {\sf Relationship constraint} 
in \cite{KF13er} had to be changed into $\{${\sf complete}$\}$ in Figure~\ref{fig:constraints} due to the multiple inheritance of
{\sf Asymmetry} that was overlooked. This has been corrected in this version of the metamodel.

\paragraph{Non-logician readability/rendering.}

\begin{figure}[t]
\centering
   \includegraphics[width=1.0\textwidth]{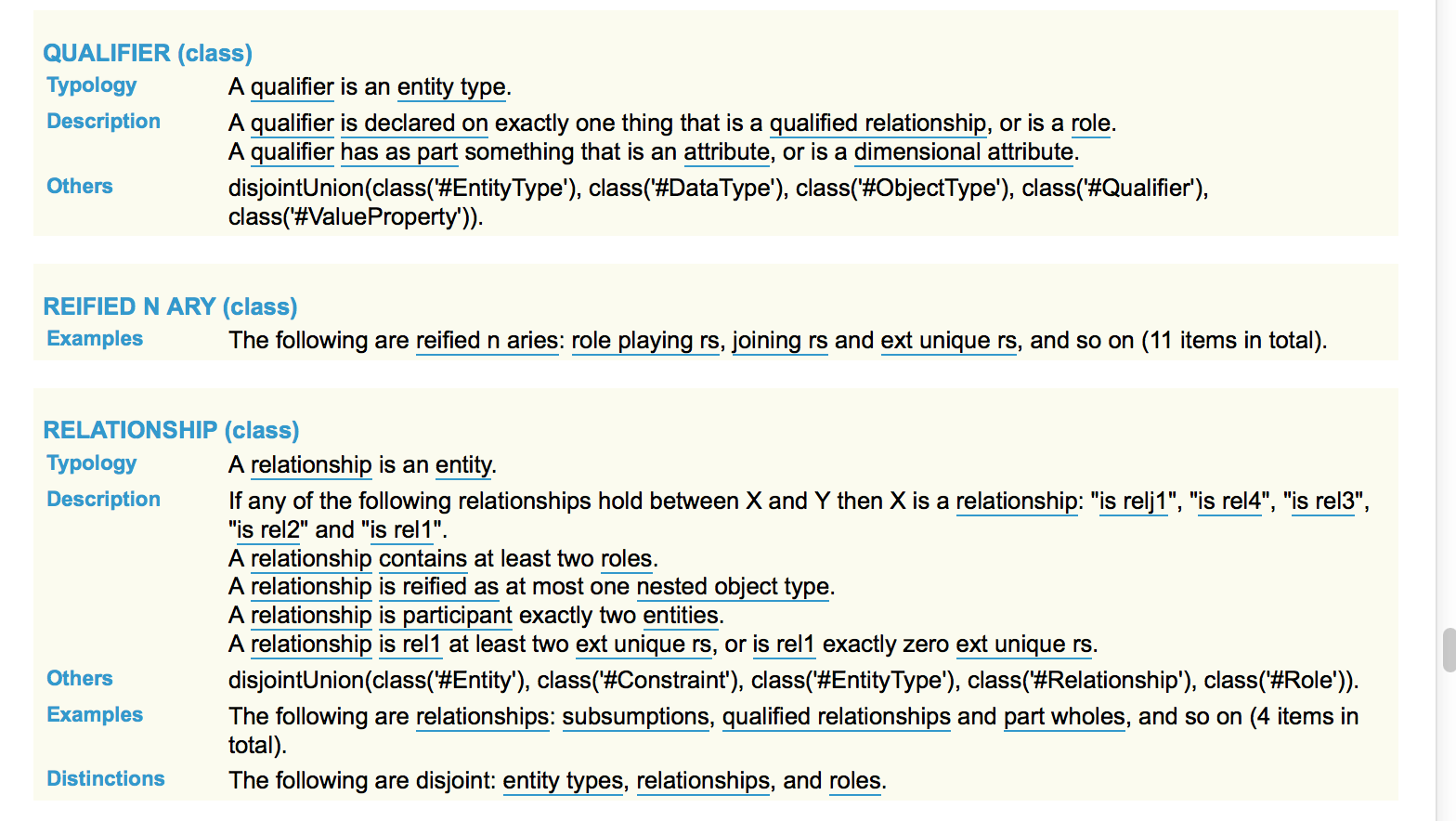}
    \caption{Section of the OWL file rendered in the SWAT NL controlled natural language (see text for details).}
    \label{fig:mmswatnl}
\end{figure}

One may expect a reader who made it this far to be familiar with at least logic or UML notation and thus have a general understanding of the contents of the metamodel and its formalisation. However, as can be observed from the previous paragraphs about the OWL version of it, there are some finer details of that logic reconstruction. Also, the FOL version of the metamodel is rather light on annotations. To facilitate readability, we have generated an online controlled natural language version (verbalization) of the OWL file, which was made with SWAT Natural Language tools \cite{Third11}. This easily navigable HTML document is available online at \url{www.meteck.org/files/ontologies/metamodelNL.html} and a section of it is shown in Figure~\ref{fig:mmswatnl}.

\section{Final remarks}
\label{sec:concl}

We have presented the FOL formalisation of the KF metamodel that serves as a unifying metamodel for the static structural entities of UML Class Diagrams (v2.4.1), ER, EER, ORM, and ORM2. It has to be noted that this indeed requires FOL. Notwithstanding, and with an eye on implementations, we also have created an OWL 2 DL version of it, noting some modelling decisions, such as a pattern for representing n-aries.

Initial use of the formalisation has been demonstrated in \cite{FK14}, where the constraints represented among the entities induce validation checking of inter-model assertions and model transformations.

\subsubsection*{Acknowledgments} 
This work is based upon research supported by the National Research Foundation of South Africa (Project UID90041) and the Argentinean Ministry of Science and Technology. Any opinion, findings and conclusions or recommendations expressed in this material are those of the author and therefore the NRF does not accept any liability in regard thereto.
 
\bibliographystyle{apalike}
\bibliography{metamodels}

\end{document}